\documentclass[10pt]{article}
\usepackage[preprint]{tmlr/tmlr}

\usepackage{macros}
\usepackage{pgfplots}
\pgfplotsset{compat=1.18}
\usepackage{pgfmath}
\usepackage{amsmath}
\usepackage{float}

\newcommand{\logit}{\mathrm{logit}}

\author{\name Jessie Finocchiaro \email finocch@bc.edu \\
      \addr Department of Computer Science\\
      Boston College
      \AND
      \name Sanket Shah\thanks{Work done while a PhD student at Harvard} \email sanketshah@g.harvard.edu \\
      \addr Independent Researcher
      \AND
      \name Milind Tambe \email tambe@g.harvard.edu\\
      \addr Department of Computer Science \\
      Harvard University 
      }
\title{Contrasting Cost-Agnostic and Cost-Sensitive Losses under Limited Model Capacity via $\mathcal H$-Consistency}

\def\month{MM}  \def\year{YYYY} \def\openreview{\url{https://openreview.net/forum?id=XXXX}}

\begin{document}

\maketitle

\begin{abstract}
There is a prevalent debate in machine learning about whether practitioners should train models to optimize a task-agnostic objective (e.g., cross entropy) or incorporate the downstream decision task into the optimization objective (e.g., \emph{weighted} cross entropy).
In ideal settings, like those with infinite data and infinite model capacity, the two approaches are statistically equivalent for the downstream decision task.
In practice, however, incorporating the decision task into model training has been shown to empirically improve task-specific performance in certain real-world scenarios.
The cause of these benefits has not been theoretically studied to date.
Focusing on the setting with limited model capacity through the model class $\mathcal H$, we establish a strict performance gap between post-processing a model learned with a cost-agnostic objective (e.g., thresholding a risk score prediction) and models learned by optimizing a cost-sensitive model without a threshold search.
In particular, we establish this gap when there is a hypothesis recovering the optimal decision boundary for the discrete task, but it does not align with the optimal cost-agnostic hypothesis, and give a simple example demonstrating the plausibility of this setting.
With assumptions that are hard to verify in practice, we demonstrate that this gap generally exists on classification datasets from the UCI repository, particularly with very simple models.
\end{abstract}

\jessie{Conjecture: regret for weighted analogs is lower than cost-agnostic?}
\claude{\textbf{Review pass: summary, updated after two follow-up revisions.} Of the original six load-bearing issues, four are now resolved. \textbf{(1)} \Cref{thm:cost-agnostic-not-H-consistent} is fixed to $\Hlin$ specifically, no longer false as quantified over generic $\H\supseteq\Hlin$. \textbf{(5)} $\Hlin$ saturates at $\pm1$ and \Cref{thm:cost-agnostic-not-H-consistent}'s admissible set is now fixed at $\psi_{1/2}$ rather than an arbitrary threshold, so it and \Cref{thm:embeddings-H-consistent} share a single, explicit $(\H,P)$ pair end-to-end. \textbf{(6)} The absolute-continuity assumption is properly scoped to exclude \Cref{ex:antisymm-counterex}'s finite support. \textbf{(4)} \Cref{thm:weighted-analogs-H-consistency} has been restructured: \Cref{assum:stationarity} now states directly that $h^\ell_w$ is the unique $L^\alpha$-optimal hypothesis over $\H$ (rather than a Bregman-generator stationarity condition), and the theorem's own proof is now a short, direct argument (parameter convergence, via \Cref{assum:convexity}+\Cref{assum:stationarity}, forces a.e.\ decision convergence, forces target-risk convergence) with no contrapositive and no remaining gaps that I'm aware of. The old calculus-based sufficient condition for \Cref{assum:stationarity} (formerly Proposition 10) has been removed entirely: it was never cited anywhere to establish the assumption for a concrete distribution, and per item (3) below, it never could have been for the running example. One item newly regresses. \textbf{(3)} \Cref{ex:binary-csc} does \emph{not} satisfy \Cref{assum:stationarity} -- confirmed independently twice, including a direct cross-check against \Cref{fig:weighted-xe-hlin}'s own $\alpha$ explaining why that figure looks correct anyway (the gap is real but small at that $\alpha$); see the note at \Cref{sec:weighted-XE-binary}. The claim that it does, in both \Cref{sec:discuss-small} and \Cref{sec:weighted-XE-binary}, needs to be rewritten, not just flagged. One item is mostly resolved. \textbf{(2)} The vertical-contours claim is now a proven equality via a symmetry argument (see the note at \Cref{ex:binary-csc}); one residual remains, since that argument needs the cost-agnostic loss to be symmetric, which \Cref{thm:cost-agnostic-not-H-consistent}'s ``any cost-agnostic surrogate'' quantifier doesn't guarantee. This revision also disambiguates $\htargetagnostic$ (now $\htargetagnostic_{\lin}$ where the $\Hlin$-restricted approximation is meant, distinct from the unconstrained population log-odds), and consolidates the empirical Embeddings/Embeddings+Softmax rows in \Cref{sec:experiments}. Remaining notation/presentation nits are collected at \Cref{tab:notation}. To hide all of these notes, set \texttt{\textbackslash Comments} to 0 in \texttt{macros.sty}.}

\section{Introduction}

In training machine learning models for discrete prediction tasks, the practitioner faces a plethora of design decisions.
One of these choices is the loss function used as a training objective: should loss functions be hand-crafted for a ``cost-sensitive'' decision task at hand, or does it suffice to post-process a model trained on a generic loss like cross entropy?
The debate about this choice is longstanding; the former camp argues that, when faced with the inevitability of imperfect prediction, optimizing a custom surrogate  will yield better predictions for the task at hand~\citep{vapnik_nature_2000,scott_calibrated_2012,ramaswamy_consistent_2016,wilder_melding_2019,elmachtoub_smart_2022,shah_decision-focused_2022}.
In contrast, the other camp argues that one can often adapt the model for the task at hand with clever post-processing while generalizing to unknown decision-makers~\citep{dawid_well-calibrated_1982,seidenfeld_calibration_1985,buja_loss_2005,reid_surrogate_2009,reid_information_2011,gopalan_omnipredictors_2022}.
As an example of such a cost-sensitive decision task, suppose the cost of a false positive is much higher than the cost of a false negative in a binary classification problem. 
For example, an IT team coping with phishing attacks will probably prefer to cope with false positives (with consequences of some genuine e-mails being flagged as spam) than false negatives (leading to data breaches arising from successful phishing attacks) in a spam detection system.
In that case, one might have a threshold $\tau \in [0,1]$ in mind where they want to flag an email as spam if a model predicts $\Pr[$spam $\mid$ email contents$] \geq \tau$.

In ``nice'' settings where perfect prediction is possible, it is known that the two approaches are statistically equivalent~\citep{frongillo_elicitation_2021}. 
However, in practice, the decision-focused learning literature has empirically improved task-specific performance by designing custom loss functions~\citep{elmachtoub_smart_2022,wilder_melding_2019}.
We examine the gap induced by loss function choice in the setting of cost-sensitive classification from a theoretical perspective.
Here, cost-sensitive classification refers to a setting where ``costs'' for different types of errors (e.g., false positives and negatives) are written as a cost matrix $\ell$. 
Our goal in this work is not to advocate for cost-sensitive surrogates over cost-agnostic ones generally, but rather to serve as a first step towards characterizing settings in which the choice of surrogate training loss affects the quality of decision-theoretic performance.
While we show positive results for cost-sensitive surrogates, we also note that cost-agnostic losses have their place.
For example, omnipredictors~\citep{gopalan_omnipredictors_2022} enable post-processing of a model's predictions for a broad class of decision-makers, which is a benefit sacrificed by selecting a cost-sensitive training loss.

We establish a gap between the performance of models trained on cost-sensitive vs agnostic losses in the setting of limited model capacity through the lens of $\H$-consistency.
$\H$-consistency examines whether learning the best model in a \emph{hypothesis class} $\H$ with respect to a training loss $L$ implies making the best decision for a cost-sensitive target problem $\ell$.
We show that, for ``small enough'' model classes $\H$ (relative to the data-generating distribution $P \in \Delta(\X \times \Y)$), there are instances where a cost-sensitive surrogate $\Ltargetaware$ is $\H$-consistent with respect to the target problem $\ell$ while a canonical cost-agnostic surrogate $\Ltargetagnostic$ is not $\H$-consistent, \emph{regardless of post-processing choices}.

Concretely, we construct a counterexample giving some intuition for our setting in \Cref{sec:example}. 
With this counterexample, we show in \Cref{sec:XE-not-H-consistent} that a cost-agnostic surrogate (e.g., cross entropy, squared error) plus post-processing are not $\H$-consistent, and can fail to perform well for the cost-sensitive target.
In contrast, we evaluate the $\H$-consistency of a class of surrogates called embeddings~\citep{finocchiaro_embedding_2024}.
These task-specific losses in~\Cref{sec:embeddings-H-consistent} include piecewise linear losses like the weighted hinge.
By ``corresponding exactly to the optimal decision for $\ell$,'' we show these losses yield $\H$-consistency guarantees in the same setting as \cref{sec:XE-not-H-consistent}. 
In \Cref{sec:weighted-XE-binary} we analyze weighted analogs of canonical cost-agnostic surrogates (e.g., Weighted Cross Entropy, Squared Error, etc.), and show that even in binary outcome settings, stronger assumptions are needed to establish $\H$-consistency.
Our theoretical results establish a strict gap between the performance of small ML models trained on cost-sensitive surrogates and those trained on cost-agnostic surrogate losses, even when the cost-agnostic losses are equipped with a threshold search to post-process model predictions.
Finally, in~\Cref{sec:experiments}, we empirically evaluate the existence of this performance gap when distributional assumptions cannot be verified.

\section{Motivating example}\label{sec:example}

\begin{figure}
\begin{minipage}{0.45\linewidth}
    \centering
    \includegraphics[width=\linewidth]{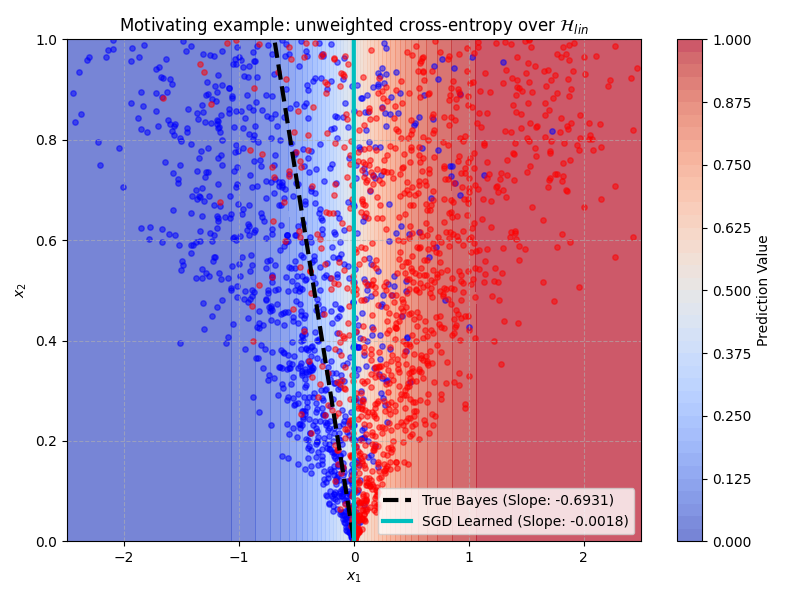}
    \caption{Cost-Agnostic Cross Entropy over linear models.} \label{fig:unweighted-xe-hlin}
\end{minipage}
\hfill
\begin{minipage}{0.45\linewidth}
    \centering
    \includegraphics[width=\linewidth]{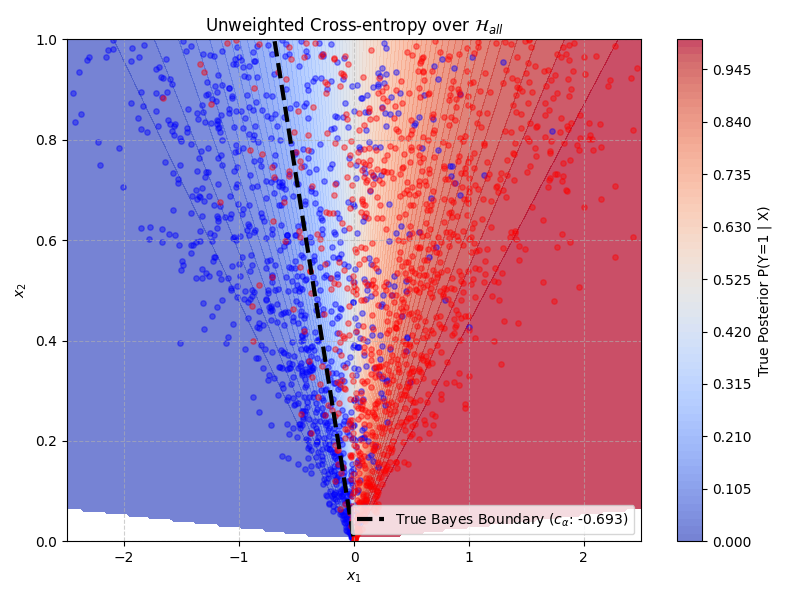}
    \caption{Unweighted Cross Entropy over all measurable functions.} \label{fig:unweighted-xe-hall}
\end{minipage}
\vfill
\begin{minipage}{0.45\linewidth}
    \centering
    \includegraphics[width=\linewidth]{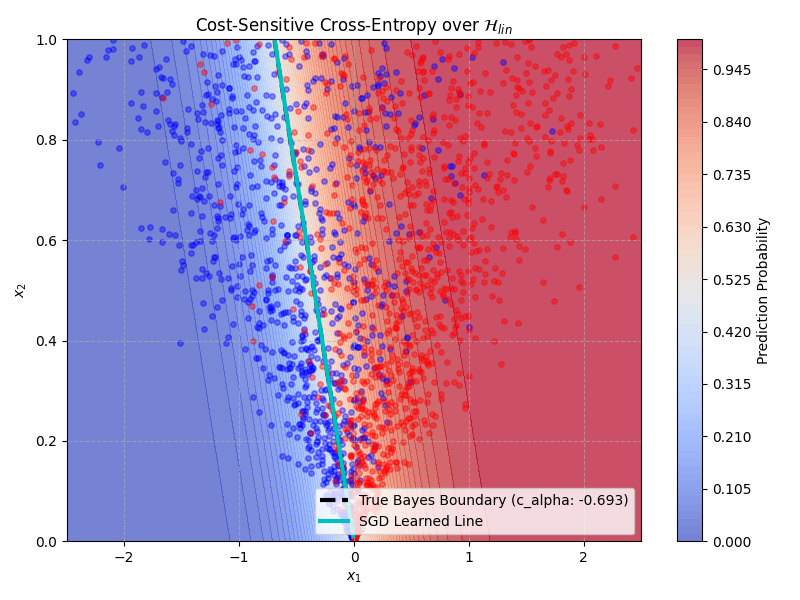}
    \caption{Cost-Sensitive Cross Entropy over linear models.} \label{fig:weighted-xe-hlin}
\end{minipage}
\hfill
\begin{minipage}{0.45\linewidth}
    \centering
    \includegraphics[width=\linewidth]{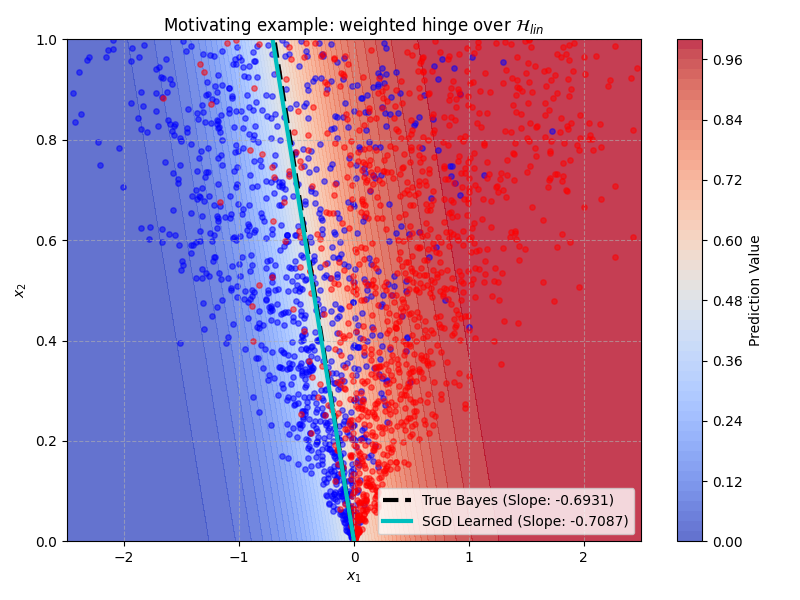}
    \caption{Weighted Hinge (An Embedding) over linear models.} \label{fig:weighted-hinge-hlin}
\end{minipage}
\caption{Contrasting the contours output by different hypotheses while varying the surrogate loss $L$ and the model capacity $\H = \{\Hlin, \Hall\}$ Notably, when optimizing over a sufficiently large class of models, the cost-agnostic surrogate yields contours that rotate, capturing the true geometry of $p(x)$.
When optimizing over $\Hlin$, the model must produce parallel contours, and is unable to rotate and capture this underlying geometry of $p(x)$. However, the cost-sensitive surrogates pick the correct slope to recover the decision boundary.}\label{fig:main-example-contours}
\end{figure}

To develop an intuition for where a performance gap between cost-sensitive and cost-agnostic models emerges, we turn to~\Cref{ex:binary-csc}, which gives a cost-sensitive binary classification problem $\ell^\alpha$ with costs in~\Cref{tab:csc-costs} and learned hypotheses visualized in \Cref{fig:main-example-contours}.
In essence, the target loss $\ell^\alpha$ imposes cost $1-\alpha$ for a false negative, and cost $\alpha$ for a false positive.
A model $h : \X \to \reals$ defines contours over the space $\X$, and the possible shapes of these contours are determined by the model class $\H$.
Predictions in $\reals$ are then post-processed into decisions in $\R \subset \mathbb{Z}$ through a link function, discussed shortly.

For our motivating example, we consider optimizing over the model class $\Hlin = \{x \mapsto \rho(w^Tx) \mid w \in \reals^2\}$, where $\rho : \reals \to [-1, 1]$ is the sign-based translation of the sigmoid $2 \sigma(z) -1$.

\begin{example}\label{ex:binary-csc}
    Consider features $\X \subseteq \reals^2$ and binary labels $\Y = \{-1,1\}$ drawn from the distribution $P$ such that 
    \begin{align*}
    Y\sim \mathbf{Bern}\left(\frac 1 2\right)~~~~
    X_2 \sim U[0,1],~~~~
    X_1 \sim \mathcal{N}(y \cdot X_2, X_2^2).
    \end{align*}
    The optimal cost-agnostic linear hypothesis over $P$ is simply $\htargetagnostic_{\lin}(x) = \rho(x_1)$\footnote{Technically, $\htargetagnostic_{\lin}(x) = \rho(c x_1)$ for some $c \neq 0$, but we leverage a degree of freedom to take $c = 1$.} because of the invariance of $P$ under $(x_1, y) \mapsto (-x_1, -y)$ for a fixed $x_2$.
    Moreover, the conditional label distribution $\Pr[Y|X = x] := p(x) = \frac{f(\frac{x_1 - x_2}{x_2})}{\left(f(\frac{x_1 - x_2}{x_2}) + f(\frac{x_1 + x_2}{x_2})\right)}$ is nonlinear in $\X$, where $f(\cdot)$ denotes the unit Gaussian PDF.
    Equivalently, $p(x) = \sigma\left(\frac{2x_1}{x_2}\right)$; despite this logistic form, no weights $w$ satisfy $\sigma(\langle w,x\rangle) = p(x)$, since $\frac{2x_1}{x_2}$ is a ratio rather than a linear function of $x$.
However, the level set $\{x : p(x) = \alpha\}$ is linear in $\X$ for all $\alpha \in (0,1)$.
\end{example}

\claude{\textbf{Mostly resolved -- answering ``J: resolved?''} For \Cref{ex:binary-csc} itself: yes. The claim is now stated as an equality via the dedicated symbol $\htargetagnostic_{\lin}$, and the justification given (``invariance of $P$ under $(x_1,y)\mapsto(-x_1,-y)$ for fixed $x_2$'') is exactly the symmetry argument I'd sketched -- $\sigma(-z)=1-\sigma(z)$ turns that invariance into $R(w_1,w_2,b)=R(w_1,-w_2,-b)$, and strict convexity in $w$ forces $w_2=b=0$. The dual-meaning bug in \Cref{sec:discuss-small} this comment used to flag is also gone: that paragraph now consistently uses $\htargetagnostic_{\lin}$ rather than reusing bare $\htargetagnostic$ for two different objects. \textbf{One genuine residual}: the symmetry argument requires the cost-agnostic loss to be \emph{symmetric} (e.g.\ cross entropy), but \Cref{thm:cost-agnostic-not-H-consistent} quantifies over ``any cost-agnostic surrogate $\Ltargetagnostic$,'' a class that also contains asymmetric proper scoring rules, for which $w_2=0$ isn't implied by this argument. Either add ``symmetric'' to that theorem's hypothesis on $\Ltargetagnostic$, or note explicitly that the argument needs revisiting for the asymmetric case.}
\jessie{Inconsistency proofs only require a single counterexample. Since Ex 1 is in the set, it suffices.}

\begin{propositionE}[][end,restate]\label{prop:example-optimal-classifiers-linear}
The $\ell^\alpha$-Bayes-optimal decision boundary for \Cref{ex:binary-csc} is linear.
\end{propositionE}
\begin{proofE}
By construction of $P$, we have 
\begin{align*}
    \Pr[Y = 1 \mid X = x] &= \frac{\Pr[X = x \mid Y = 1] \Pr[Y = 1]}{\Pr[X = x]} \\
    &=\frac{f(\frac{x_1 - x_2}{x_2}) \frac 1 2}{\frac{1}{2}\left(f(\frac{x_1 - x_2}{x_2}) + f(\frac{x_1 + x_2}{x_2})\right)} \\
    &=\frac{f(\frac{x_1 - x_2}{x_2})}{\left(f(\frac{x_1 - x_2}{x_2}) + f(\frac{x_1 + x_2}{x_2})\right)} ~.~
\end{align*}
where $f(\cdot)$ is the pdf of the unit Gaussian.

Therefore, with cost matrix $C(r,y) = \begin{bmatrix} 0 & 1-\alpha \\ \alpha & 0 \end{bmatrix}$, given any $x$, we can ask when the expected \emph{cost} of predicting when $r = 1$ is better than $r = -1$

\begin{align*}
    \sum_{y \in \{-1,1\}} \Pr[Y = y \mid X = x] C(1, y) &\leq \sum_{y \in \{-1,1\}} \Pr[Y = y \mid X = x] C(-1, y) \\
    \alpha \Pr[Y = -1 \mid X = x] &\leq (1- \alpha) \Pr[Y = 1 \mid X = x] \\
    \alpha (1 - \Pr[Y = 1 \mid X = x]) &\leq (1- \alpha) \Pr[Y = 1 \mid X = x] \\
    \alpha &\leq \Pr[Y = 1 \mid X = x] \\
    \alpha &\leq \frac{f(\frac{x_1 - x_2}{x_2})}{\left(f(\frac{x_1 - x_2}{x_2}) + f(\frac{x_1 + x_2}{x_2})\right)} \\
    \alpha &\leq  \frac{\exp(-\frac 1 2(\frac{x_1 - x_2}{x_2})^2)}{\exp(-\frac 1 2(\frac{x_1 - x_2}{x_2})^2) + \exp(-\frac 1 2(\frac{x_1 + x_2}{x_2})^2)} \\
    \alpha &\leq \frac{\exp(-\frac 1 2(\frac{x_1 - x_2}{x_2})^2)}{\exp(-\frac 1 2(\frac{x_1 - x_2}{x_2})^2) + \exp(-\frac 1 2(\frac{x_1 + x_2}{x_2})^2)} \\
    \frac{1}{\alpha} &\geq \frac{\exp(A) + \exp(B)}{\exp(A)} \qquad A = -\frac 1 2(\frac{x_1 - x_2}{x_2})^2, B = -\frac 1 2(\frac{x_1 + x_2}{x_2})^2 \\
    \frac{1}{\alpha} &\geq 1 + \frac{\exp(B)}{\exp(A)} \\
    \frac{1}{\alpha} &\geq 1 + \exp(B-A) \\
    \frac{1}{\alpha} &\geq 1 + \exp(\frac{-2x_1}{x_2}) \\
    \log\left(\frac{1}{\alpha}-1\right) &\geq \frac{-2x_1}{x_2}  \\
    x_2 \log\left(\frac{1}{\alpha}-1\right) &\geq -2x_1 \\
    \frac{x_2}{2} \log\left(\frac{\alpha}{1-\alpha}\right) &\leq x_1
\end{align*}
\end{proofE}

Example \ref{ex:binary-csc} has a few nice properties: first, the optimal cost-sensitive classifier is linear (Proposition~\ref{prop:example-optimal-classifiers-linear}).
Moreover, the optimal cost-sensitive and cost-agnostic linear classifiers are \emph{different} from each other.
Namely, the optimal linear cost-agnostic classifier is $\htargetagnostic_{\lin}(x) = \rho(x_1)$ (\Cref{fig:unweighted-xe-hlin}, which yields vertical contours), while the optimal target classifier is $h^\alpha(x) = \rho\left(x_1 + \frac{1}{2} \log\left(\frac{1-\alpha}{\alpha}\right) x_2 \right)$ (\Cref{fig:weighted-xe-hlin,fig:weighted-hinge-hlin}).

Now, we can incorporate post-processing by choosing a link $\psi_\tau(u) = \sign(u - \tau)$ parameterized by $\tau$, and modify these thresholded classifiers by $\psi_\tau \circ \htargetagnostic_{\lin}(x) := \sign(\rho(x_1) - \tau)$ to enable a threshold search, as is common in practice. 
Observe that the threshold $\tau$ permits shifting of the bias along the contours, but not the slope of the learned model. 
Therefore the optimal linear classifiers $\htargetagnostic_{\lin}(\cdot)$ and $h^\alpha(\cdot)$ have different contours for $\alpha \neq \frac 1 2$, and thresholding cannot recover the non-vertical slope of $h^\alpha$.
This implies some cost-sensitive decisions will be made erroneously.

\subsection{Characterizing Limited Model Capacity}\label{sec:discuss-small}
To clarify what we mean by ``small'' model classes or ``limited model capacity'' throughout our paper, we lean on \Cref{ex:binary-csc}. 
The notion of ``small enough'' in our work leverages the separation that the $\ell$-optimal model is contained in $\H$, but the $\Ltargetagnostic$-optimal classifier is not contained in $\H$.

The general approach takes the following steps: (1) given a loss $L$, find its Bayes optimal classifier $h^L(x) \equiv \Gamma(p(x))$, through the property $\Gamma$ elicited by $L$ (\Cref{subsec:surrogates}). 
(2) Diagnose if $h^L \in \mathcal H$, and 
(3a) if $h^L \in \mathcal H$ and $h^L = h^\ell$, \citet[Theorem 2.8]{steinwart_how_2007} proves that $\H$-calibration and $\H$-consistency are equivalent. 
We can proceed with a pointwise analysis of the optimal decision via a sufficient condition called $\H$-calibration (\Cref{sec:H-calibration}). 
(3b) If $h^L \in \H$ but $h^L \neq h^\ell$, the path to $\H$-consistency still might hold through distributional assumptions (e.g., \Cref{assum:convexity} and \Cref{assum:stationarity}).
(3c) If $h^L \not \in \H$, analysis becomes complex, but previous results have bounded consistency rates in terms of the \emph{minimizability gap} and proceeded with analysis in the manner of (3a)~\citep{mao2025principledmetrics,mao_h-consistency_2024}, folding in the minimizability gap as an additive regret term.
While some literature does proceed by assuming the conditions of Step (3a), we generally do not assume $h^L = h^\ell$, instead establishing that the condition holds for a class of piecewise linear surrogates.

To gain intuition for this approach, consider \Cref{ex:binary-csc}, with the class $\Hlin$: the $\Hlin$-optimal classifier for the \emph{unweighted cross entropy} is $\htargetagnostic_{\lin}(x) = \rho(x_1)$, which is independent of $\alpha$. 
This hypothesis $\htargetagnostic_{\lin}$ stands in contrast to the $\ell^\alpha$-optimal decision boundary with slope that is a function of $\alpha$.
With $\htargetagnostic_{\lin}(x) = \rho(x_1)$, we claim there is no threshold $\tau$ such that $\rho(x_1) \geq 2 \tau- 1 \iff \sigma(\frac{2x_1}{x_2}) \geq \alpha$, and therefore we cannot recover optimal decisions via $\htargetagnostic$.

In contrast, consider the weighted hinge: $\Ltargetaware(r,y) =  [(1-\alpha)\mathbf{1}(y = 1) + (\alpha)\mathbf{1}(y = -1)]\max(1 - ry, 0)$.
The $\Ltargetaware$-optimal for weighted hinge in this case is $\htargetaware(x) = \sign(p(x) - \alpha)$. 
Consider the linear score $\htargetaware_{\lin}(x) = 2w_1 x_1 - \logit(\alpha) w_1 x_2 = 2w_1 x_1 - \log(\frac{\alpha}{1-\alpha}) w_1 x_2$, so that $\rho(\htargetaware_{\lin}(x)) \in \Hlin$ (\Cref{subsec:H-consistency}). 
With some arithmetic manipulation, we obtain $\rho(\htargetaware_{\lin}(x)) \to \htargetaware(x)$ pointwise a.e. as $w_1 \to \infty$.
Moreover, $\htargetaware_{\lin}(x) \geq 0 \iff \sigma(\frac{2x_1}{x_2}) \geq \alpha \iff p(x) - \alpha \geq 0$, and therefore taking the sign of $\htargetaware_{\lin}(x)$ yields the $\Ltargetaware$-optimal decisions for this cost-sensitive problem. 
The additional strength of this weighted hinge is that $\sign(\htargetaware_{\lin}(x))$ coincides with the optimal decision for the cost matrix $\ell^\alpha$.

\claude{\textbf{Core fix confirmed in place, one action item still open.} The original claim ($\htargetaware_{\lin}(x) \to \htargetaware(x)$ pointwise as $w_1\to\infty$) was false: that score \emph{diverges}. Worse, the same defect recurred structurally -- \Cref{thm:embeddings-H-consistent} needs $\H$ approximately closed under the embedding link, i.e.\ sequences in $\H$ approaching the embedding points $\pm1$, and the former $\Hlin = \{\sigma(\langle w,x\rangle)\}$ saturates at $\{0,1\}$ while $\Hlin' = \{\langle w,x\rangle+b\}$ diverges. Redefining $\Hlin$ to saturate at $\pm1$ via $\rho = 2\sigma-1$ (now in \Cref{subsec:H-consistency}) fixes both at once, and the running example now correctly uses $\rho(\htargetaware_{lin}(x)) \to \htargetaware(x)$ a.e.\ -- confirmed this is in place. \textbf{Sidestepped rather than resolved}: the ``Model Class Choice Is Important'' bullet (which attributed the bad Synthetic number to training under $\Hlin'$) has been removed, and \Cref{tab:results}'s Synthetic ``Embeddings'' row now reports the softmax parameterization (0.353, beating both cross entropy rows) instead of the old plain-$\Hlin'$ number (0.428, now commented out but still in the source). That's a real improvement for the reader -- the confusing worse-than-cost-agnostic number is no longer presented as ``the'' embeddings result -- but it doesn't answer the original question, since Embeddings+Softmax is a genuinely different parameterization from $\Hlin$ (see the note in \Cref{app:additional-exp}), not the corrected saturating $\Hlin$ itself. Whether embeddings trained on literal $\Hlin=\{\rho(w^\top x)\}$ now performs well (as the corrected theory predicts) is still untested in what's reported.}
\jessie{I removed the ``Model Class Choice is Important'' bullet. While it is important, I think it detracts from the main message. But using $\sigma$ vs $\rho$ in model class definition should not matter from an empirical perspective. It's the same up to relabeling.}

Finally, consider weighted cross entropy: $L^\alpha(r,1) = -(1-\alpha)\log(r)$ and $L^\alpha(r,-1) = -\alpha \log(1-r)$ obtains a Bayes optimal classifier $h^\alpha(x) = \frac{(1-\alpha) \sigma(\frac{2x_1}{x_2})}{\alpha + \sigma(\frac{2x_1}{x_2}) - 2 \alpha \sigma(\frac{2x_1}{x_2})}$. 
Observe that $h^\alpha\not \in \Hlin$, but  $h^\alpha(x)\geq \frac 1 2$ if and only if $\sigma(\frac{2x_1}{x_2}) \geq \alpha$, which would allow us to recover the optimal decision, if optimizing $L^\alpha$ over $\Hlin$ yields $h^\alpha$.

\section{Background and Related Work}

We are primarily concerned with minimizing expected loss, where data is sampled from a distribution $P \in \Delta(\X \times \Y)$ over features $\X$ and labels $\Y$, with $|\Y|$ finite.
We often denote with $p \in \simplex$ an arbitrary distribution over labels $\Y$.
When $P$ is clear from context, we often let $p(x) := \Pr_P[Y | X = x]$ denote the conditional distribution over labels given a feature vector $x$.
In binary settings, where $\Y = \{-1,1\}$, we abuse notation and write $p(x) = \Pr_P[Y = 1|X= x]$ to denote the probability of a positive label.
Let $P_X$ denote the marginal distribution of $P$ over $\X$.
Throughout, we assume $P_X$ is absolutely continuous with respect to Lebesgue measure on $\X$ when $\X$ is a linear subspace of $\reals^k$; that is, $P_X(S) = 0$ for any Lebesgue-null $S \subseteq \X$.
This lets us treat lower-dimensional sets (e.g., decision boundaries, ties between reports) as $P_X$-null without further comment.

\subsection{Decision and surrogate losses}
In classification-like (discrete) prediction tasks, we can formalize the costs of a model's decisions through a discrete loss $\ell : \R \times \Y \to \reals_+$ mapping a finite set of \emph{decisions}, or \emph{reports}, $\R$ to a real-valued error when compared against the \emph{true outcome} in $\Y$.
Equivalently, we often denote $\ell$ by a \emph{cost matrix} $\mathbf{\ell} \in \reals^{|\R| \times |\Y|}$, where $\ell(r,y) = \mathbf{\ell}_{r,y}$.
This approach encapsulates broader classes of problems such as ranking~\citep{ramaswamy_convex_2016} and classification with rejection~\citep{ramaswamy_consistent_2016}, though our primary motivation is cost-sensitive classification.
\Cref{tab:csc-costs} gives one example of such a cost-sensitive discrete loss $\ell^\alpha$ for the binary report and label setting.

\begin{table}[h]
\centering

\begin{tabular}{cr|cc}
     \multicolumn{2}{c|}{\multirow{2}{*}{\makecell{Cost\\Matrix}}} & \multicolumn{2}{c}{\color{black!50} True label} \\
     &  & -1 & 1 \\
    \midrule
    \multirow{2}{*}{\rotatebox[origin=c]{90}{\color{black!50} Pred}} & -1 & 0 & $1-\alpha$ \\
     & 1 & $\alpha$ & 0
    \end{tabular}
\caption{Cost-sensitive classification loss $\ell^\alpha$ for binary outcomes. We normalize the loss so it can be represented with a single parameter, as done by \citet{elkan_foundations_2001,scott_calibrated_2012}. This parameterization precludes application of results from \citet{mao2025principledmetrics}.}
    \label{tab:csc-costs}
\end{table}

Unfortunately, discrete losses are often intractable to optimize directly~\citep{ben2003difficulty}, which necessitates we instead optimize \emph{surrogate losses} $L : \reals^d \times \Y \to \reals_+$.
These surrogates are then \emph{post-processed} via a link function $\psi : \reals^d \to \R$ which maps real-valued surrogate predictions to discrete decisions.
Canonical links for classification take the argmax of scores corresponding to each class, and cost-sensitive analogs might take the argmax of a linear transformation of these scores.
For example, the ``ideal'' model $h^* : \X \to \reals^{|\Y|}$ might have $h^*(x) \approx p(x) \in \reals^{|\Y|}$ for all $x$. 
In this case, using a link such as $\psi(u) = \argmax_y u_y$ roughly corresponds to learning the most likely label conditioned on the feature vector $x$, which minimizes the 0-1 cost matrix $\ell = \ones \ones^T - I$.
If the cost matrix was a generic loss matrix $\ell$, the conditional label predictor $h^*(x) = p(x)$ can always be post-processed to obtain optimal decisions for $\ell$, as implied by results in property elicitation~\citep{lambert_eliciting_2009,lambert_elicitation_2018,frongillo_elicitation_2021}.
We now discuss the surrogate losses that we study in more detail in \Cref{subsec:surrogates}, specifically discussing ``cost-sensitive'' surrogates through the lens of property elicitation.

\subsection{Cost-sensitive vs cost-agnostic surrogates, property elicitation}\label{subsec:surrogates}

To state explicitly the relationship of a cost-agnostic surrogate and the cost-sensitive surrogate, we turn to \emph{property elicitation}.

\begin{definition}[(Indirect) Property Elicitation]
Fix a set $\P \subseteq \simplex$.
A property $\Gamma: \P \to \R$ is a function mapping probability distributions over labels in $\Y$ to reports in $\R$. 
A loss $L : \R \times \Y \to \reals_+$  \emph{elicits} $\Gamma$ over $\P$ if, for all $p \in \P$, we have
\begin{align*}
\Gamma(p) \in \argmin_{r \in \R} \E_{Y\sim p} L(r, Y)~.
\end{align*}
Moreover, we say $L$ which directly elicits $\Gamma$, \emph{indirectly elicits} $\gamma : \P \to \hat \R$ if there is a link $\psi : \R \to \hat \R$ such that $\gamma = \psi \circ \Gamma$.
If $\P = \simplex$, we omit ``over $\P$.''
\end{definition}

In what is to come, consider the property $\gamma$ directly elicited by our cost matrix $\ell$.
We reason about cost-agnostic surrogates as those which indirectly elicit $\gamma$ by using $\gamma$ as the link.
That is, the cost-agnostic surrogate directly elicits the identity $p$, which is justified by the observation that since $\gamma$ is a function of $p$, one can simply post-process $p$ to obtain $\gamma(p)$.
In contrast, cost-sensitive surrogates also indirectly elicit $\gamma$, but through a different link such as a relabeling of reports.
Importantly, the cost-agnostic surrogate and cost-sensitive surrogates both indirectly elicit the property $\gamma$.

\subsubsection{Cost-Agnostic Surrogates: Proper Scoring Rules}
One special class of cost-agnostic surrogates to consider is \emph{proper scoring rules}~\citep{savage_elicitation_1971,gneiting_strictly_2007,brier_verification_1950}.
These are generally regarded as loss functions that ideally lead to the predictor of the conditional label distribution $\htargetagnostic(x) = p(x)$ in ideal conditions. 

\begin{definition}[Proper scoring rule]
   A loss $L : \simplex \times \Y \to \reals_+$ is a \emph{proper scoring rule} if, for all $p,q \in \simplex$, we have
   \begin{align*}
       \E_{Y \sim p}L(p,Y) \leq \E_{Y \sim p}L(q, Y)~.
   \end{align*} 
   Moreover, $L$ is \emph{strictly proper} if the inequality is strict for all $p \neq q$.
\end{definition}
Notably, all proper scoring rules (attaining their infimum) directly elicit the \emph{identity property} $\Gamma(p) = p$.
If a loss function directly elicits the identity, we call it \emph{cost-agnostic}.
Examples of proper scoring rules include cross entropy, Brier score, and the Convolutional Fenchel-Young loss~\cite{cao2026establishing}.

\subsubsection{Cost-Sensitive surrogates}

We discuss ``cost-sensitive surrogates'' as those which indirectly elicit a property of interest $\gamma$, but do not directly elicit the identity $\Gamma(p) = p$.
In particular, we consider two primary classes of cost-sensitive surrogates: \emph{embeddings} and \emph{weighted analogs} of proper scoring rules (for binary outcome settings).
While other cost-sensitive surrogates pervade the literature~\citep{ramaswamy2015convex,ramaswamy_convex_2016,yu_lovasz_2018,zhang_bayes_2020,khurana2025consistency,zhang2026exponentialconvexcalibrationdimension, zhang2026exactrankconvexcalibration}, they often are constructed without the desired statistical guarantees, even with unlimited model capacity, or are arduously constructed with consistency guarantees for a specific $\ell$.

\paragraph{Embeddings}
In \Cref{sec:embeddings-H-consistent}, we study a class of piecewise linear and convex surrogates called \emph{embeddings}~\citep{finocchiaro_embedding_2024}, which are constructed directly from a given cost matrix $\ell$ and yield a link function $\psi$ to round predictions $u \in \reals^d$ to discrete reports in $\R \subset \mathbb{Z}^d$.
The optimal reports minimizing these surrogates are ``directly learning'' the optimal decision for the loss matrix $\ell$.
For example, the hinge loss $L(u,y) = (1 - uy)_+$ is an embedding with labels $\Y = \{-1,1\}$ for the 0-1 cost matrix.
Unlike cross entropy, the minimizers of the expected (in the outcome) hinge are precisely at the reports $u^* \in \{-1,1\}$ for (almost all) distributions on $p \in \simplex$.
Informally, these piecewise linear cost-sensitive surrogates sacrifice a level of granularity in predictions present in cross entropy which can be overly burdensome in limited-model settings.
Instead of the optimal classifier returning $p(x)$ like proper scoring rules, we have the optimal classifier $\htargetaware(x) = \gamma(p(x)) = \sign\left(p(x) - \frac 1 2\right)$, minimizing the expected discrete loss, up to a relabeling of the minimizer by the link.
Importantly, in constructing these surrogates, \citeauthor{finocchiaro_embedding_2024} also give a link construction, eliminating the need to conduct a threshold search for a link function.

\paragraph{Weighted analogs of proper scoring rules}
We consider a loss function $L^\alpha$ to be an $\vec \alpha$-weighted analog of a cost-agnostic surrogate $L$ if, for all $u \in \reals^d, y \in \Y$, we can write $L^{\alpha}(u,y) = \alpha_y L(u,y)$.
This permits cost-agnostic losses to be scaled by the outcome.
For example, in \Cref{tab:csc-costs}, we parameterize a false positive with weight $\alpha$ and false negative with weight $1-\alpha$; one popular weighted analog in this cost-sensitive binary classification task is the \emph{weighted} cross entropy, with $L^\alpha(u,1) = (1-\alpha) L(u,1)$ and $L^\alpha(u,-1) = \alpha L(u,-1)$.
With binary outcome and report matrices, these weighted analogs can be parameterized without loss of generality by a single scalar, as in \Cref{tab:csc-costs}.
We let the binary case be our main focus, as generalizing results for these weighted analogs to multiclass settings is not straightforward as the number of parameters grows in $O(|\R| |\Y|)$, though it poses an interesting line of future work.

\subsection{$\H$-consistency}\label{subsec:H-consistency}
We seek guarantees on surrogate and link pairs $(L,\psi)$ which imply optimizing a surrogate loss $L$ over a model class $\H$ and applying the link function $\psi$ to obtain discrete decisions yields the same discrete decisions as if one had optimized $\ell$ directly.
This guarantee is formalized by the notion of $\H$-consistency in Definition \ref{def:H-consistent}.
We let $\H$ denote an abstract model class, and sometimes refer to $\Hall^d := \{h : \X \to \reals^d \text{ measurable}\}$ denoting the set of all measurable hypotheses, and $\Hlin^d := \{h : \X \to \reals^d | \exists w \text{ s.t. } h(x) = \rho(\langle w,x \rangle) \}$ be the set of linear functions with a saturating activation $\rho$ applied.
When clear from context, we omit $d$.
We write $\rho(z) := 2\sigma(z) - 1$ for this activation, reserving $\sigma$ for the logistic function itself, so that $\Hlin$ has range $(-1,1)$ rather than $(0,1)$.
The purpose of this choice is that the values at which members of $\Hlin$ saturate then coincide with the embedded reports of the cost-sensitive surrogates studied in \Cref{sec:embeddings-H-consistent}.
In the binary setting we study, nothing depends on the particular $\rho$, and we fix $\rho = 2\sigma - 1$ only for concreteness.
Cost-agnostic surrogates and their weighted analogs, whose arguments are probabilities, are instead applied to the \emph{probability coding} $q(u) := \frac{u+1}{2}$.
Since $q(\rho(z)) = \sigma(z)$, probability-coded expressions retain their usual logistic form; and because $q$ is a fixed bijection, the two codings realize the same decision boundaries and differ only by a relabeling of the elicited property.

Agnostic learning aims to find a model $h \in \H$ such that the expected loss of $h$ is close to the best expected discrete loss possible in $\H$.
However, given the computational hardness of directly optimizing discrete losses, $\H$-consistency aims to guarantee that optimizing a \emph{surrogate} loss $L : \reals^d \times \Y \to \reals_+$ and applying a \emph{link} function $\psi : \reals^d \to \R$ yields the desired agnostic learning guarantee.
Without loss of generality, we assume $\R \subset \mathbb{Z}^d$ and reports in $\mathbb{Z}^d$ can be mapped to discrete classes as appropriate.

\begin{definition}[$\H$-consistency]\label{def:H-consistent}
    A surrogate $L : \reals^d \times \Y \to \reals_+$ and link $\psi : \reals^d \to \R$ pair $(L, \psi)$ is $\H$-consistent with respect to a target loss $\ell : \R \times \Y \to \reals_+$ over a set of distributions $\Q \subseteq \Delta(\X \times \Y)$ if, for all sequences $\{h_n\} \in \H$ and $P \in \Q$, we have
    \begin{multline}
        \resizebox{0.5\textwidth}{!}{$\underset{\text{approaching the best-in-class surrogate prediction}}{\underbrace{\E_{(X,Y) \sim P}L(h_n(X), Y) \to \inf_{h^* \in \H} \E_{(X,Y) \sim P}L(h^*(X), Y)}} \implies$
        }\resizebox{0.5\textwidth}{!}{$\underset{\text{approaching the best-in-class target loss by linking surrogate predictions}}{\underbrace{\E_{(X,Y) \sim P}\ell(\psi(h_n(X)), Y) \to \inf_{h^* \in \H} \E_{(X,Y) \sim P}\ell(\psi(h^*(X)), Y)}}$ 
        }\label{eq:H-consistency}
    \end{multline}
\end{definition}

Unfortunately, analyzing for which surrogates $\H$-consistency holds can be technically challenging, which has historically led researchers to study $\H$-calibration as a ``pointwise proxy'' (\Cref{def:H-calibration}).
While the two can be equivalent (e.g., when $\H = \Hall$~\citep{bartlett_convexity_2006}), their equivalence is characterized by \citet[Theorem 2.8]{steinwart_how_2007} when $h^L \in \H$ and $h^L = h^\ell$.
Notably, the conditions of \citet[Theorem 2.8]{steinwart_how_2007} do not generally hold in our setting, which we discuss shortly.

\paragraph{Distributional assumptions}
In the literature, two common distributional assumptions are made: \textit{realizability} and \textit{minimizability}.
If a loss $L$ is $P$-realizable under $\H$, then there exists a hypothesis $h^* \in \H$ such that $\E_{(X,Y)\sim P}L(h^*(X),Y) = 0$.
Notably, this means that $h^*$ yields a deterministic labeling over outcomes, and there is no outcome uncertainty.
Even under these strict distributional assumptions, \citet{long_consistency_2013,kuznetsov2014multi} observe a gap between $\H$-consistency and $\Hall$-consistency, where (a) the cross entropy loss, which is $\Hall$-consistent, is not $\Hlin$-consistent for multiclass classification, and (b) the Crammer-Singer loss, which is not $\Hall$-consistent, is realizable $\H$-consistent for multiclass classification.
This second result does not extend beyond the realizable setting.

Less stringent is the assumption that a loss $L$ is \emph{$P$-minimizable} over $\H$, meaning that a \emph{Bayes optimal} classifier is in class.
However, its error may not be $0$ as in the realizable setting due to inherent uncertainty on the labels.

\begin{definition}[$P$-minimizability, Bayes Optimal]\label{def:P-minimizable}
    For a given distribution $P \in \Delta(\X \times \Y)$, a loss $L: \reals^d\times \Y\to \reals_+$ is $P$-minimizable over $\H$ if for all $\epsilon > 0$, there exists an $h_\epsilon \in \H$ such that $\E_{Y\sim p(x)}L(h_\epsilon(x), Y)<  \inf_{u \in \reals^d}\E_{Y\sim p(x)}L(u, Y) + \epsilon$ for all $x \in \X$.
    If $\{h_\epsilon\} \to h^L$ as $\epsilon \to 0$, we call $h^L$ the \emph{$L$-Bayes optimal}.
\end{definition}
Sometimes, to simplify exposition, we denote $\htargetagnostic$ as the $\Ltargetagnostic$-Bayes optimal, and similarly, $\htargetaware$ as the $\Ltargetaware$-Bayes optimal hypotheses.
Under mild assumptions on a loss $L$ (namely, that it attains its infimum for all distributions) eliciting $\Gamma$, we have $h^L(x) \equiv \Gamma(p(x))$, making property elicitation a nice lens to study these optimal hypotheses.

$P$-minimizability of both $L$ and $\ell$ tends to be the most common distributional assumption~(e.g., \citep{bao_calibrated_2021,awasthi2021calibration}), as it often simplifies analysis to a ``pointwise optimality'' in $\H$-calibration, defined in \Cref{sec:H-calibration}.

One of our primary assumptions is $P$-minimizability of the \emph{post-processed} $\ell$-Bayes optimal hypothesis.
Crucially, \Cref{assum:ell-P-minimizable} does \emph{not} imply that the optimized surrogate $L$ is $P$-minimizable over $\H$, which is assumed by \citet[Theorem 2.8]{steinwart_how_2007} to obtain the equivalence of $\H$-calibration and $\H$-consistency.
In our work, the two are not equivalent, though $\H$-calibration implies $\H$-consistency.

\begin{assumption}\label{assum:ell-P-minimizable}
There exists a link function $\psi:\reals^d \to \R$ such that $\ell$ is $P$-minimizable over $\psi \circ \H$.
\end{assumption}

\Cref{assum:ell-P-minimizable} is equivalent to the assumption that the \emph{minimizability gap} $\inf_{h \in \H} \E_{X,Y} \ell(\psi(h(X)), Y) - \E_X \inf_{h \in \H} \E_{Y \sim p(x)} \ell(\psi(h(x)), Y)$ of $\ell$ is $0$.
Notably, we are not making this assumption for the surrogate $L$; the crux of the proof of \Cref{thm:embeddings-H-consistent} is establishing a minimizability gap of $0$ for a specific class of piecewise linear surrogates (\Cref{thm:embedding-elicits-same}).

Our second primary assumption is a notion of \emph{approximate closure} of a model class $\H$, which will later enable us to characterize model classes whose hypotheses can be linked to obtain the optimal decision boundary assumed to exist in \Cref{assum:ell-P-minimizable}.

\begin{assumption}[Approximate closure of $\H$, with dominated witnesses]\label{assum:H-closed-under-phi}
Fix a surrogate loss $L : \reals^d \times \Y \to \reals_+$ and the link $\psi : \reals^d \to \R \subset \mathbb{Z}^d$ of \Cref{assum:ell-P-minimizable}.
For every $h \in \H$ there is a sequence $\{h_n\} \subseteq \H$, which we call a \emph{witnessing sequence} for $h$, and a dominating function $b_h \in \mathcal L^1(P_X)$, such that $h_n(x) \to \psi(h(x))$ for almost all $x \in \X$ and
\begin{align*}
\left|\E_{Y\sim p(x)}L(h_n(x),Y) - \E_{Y\sim p(x)}L(\psi(h(x)),Y)\right| \leq b_h(x)
\end{align*}
for all $n$ and all $x \in \X$.
When only the convergence $h_n(x) \to \psi(h(x))$ is at issue, we say $\H$ is \emph{approximately closed under $\psi$}.
\end{assumption}

\claude{\textbf{Confirmed resolved}: the a.e.\ relaxation is in place, the domination condition is now stated directly here (not just imported from a footnote), and the unused ``family $\Psi=\{\psi_i\}$'' clause is gone -- consistent with \Cref{thm:weighted-analogs-H-consistency}'s first line, which references $\{\psi,\psi'\}$, still being commented out. \textbf{Confirmed resolved}: $\Hlin$ now saturates at $\pm1$ (\Cref{subsec:H-consistency}), matching the embedding points, so this assumption is witnessable for the running example where it previously failed. \textbf{Still open, unchanged}: \textbf{typing}. $\psi$ maps $\reals^d \to \R$ while $h_n$ maps into $\reals^d$, so ``$h_n(x) \to \psi(h(x))$'' does not type-check -- it needs $\varphi \circ \psi$, with $\varphi$ the embedding of $\R$ into $\reals^d$ from \Cref{def:embeddings}. The same substitution is needed in the domination display.}
\jessie{Updated, but need to make sure change is consistently applied}

We consider approximate closure of $\H$ to be a mild condition, as a hypothesis class might be appended to include post-processing $\H' = \psi \circ \H$.
This approach is taken by \citet{zhang_bayes_2020}, who study $\H'$-consistency when optimizing over $\H$.
However, \citet{zhang_bayes_2020} focus on the realizable setting, and do not study task-specific surrogates.

\subsection{Revisiting the main question}
We now state our main question: \emph{can we identify sufficient conditions for the existence of a strict gap between the performance of models trained on \hlc[myBlue]{cost-sensitive surrogate} loss functions and models trained on cost-agnostic surrogates combined with some \hlc[myPurple]{``clever'' post-processing}?}
Our results establish this gap in the setting of \hlc[myYellow]{small} models.

As discussed earlier, \hlc[myBlue]{cost-sensitive surrogates} refer to continuous loss functions $L : \reals^d \times \Y \to \reals_+$ designed with the intention of emulating a discrete target problem $\ell : \R \times \Y \to \reals_+$.
Formally, we say cost-sensitive surrogates indirectly elicit the property $\gamma$ elicited by $\ell$, and do not directly elicit the identity $\Gamma(p) = p$.
The cost-sensitive surrogates we use as a case study in \Cref{sec:embeddings-H-consistent} are piecewise linear and convex, and constructed directly from the discrete target $\ell$.
In the cost-agnostic surrogate setting, we formalize \hlc[myPurple]{clever post-processing} by a search over possible link functions $\psi : \reals^d \to \R$ mapping surrogate predictions, such as risk scores, to discrete decisions, such as positive or negative classifications.
The most common example of this clever post-processing in binary outcome settings is through a threshold search to establish decision boundaries $\psi_\tau(h(x)) = \sign(h(x) - \tau)$. 
However, this threshold search is often difficult to scale beyond binary classification as the search space grows exponentially in the number of classes.

Our results show that when models are \hlc[myYellow]{small}, there is a strict gap between these two approaches. 
In particular, we assume the model class $\H$ is rich enough to contain the $\ell$-Bayes optimal decision boundary, but not so rich as to contain the $\Ltargetagnostic$-Bayes optimal hypothesis predicting the conditional label distribution $p(x)$ for all covariates $x \in \X$.

Our approach merges the perspectives of two fields: the work on surrogate loss design and the work on $\H$-consistency.
The surrogate loss design literature generally assumes that the model class $\H$ is rich enough for any measurable hypothesis to be learned, ignoring the relationship between model class and learned hypothesis (e.g., \citet{reid_information_2011,reid_surrogate_2009,buja_loss_2005,bao_proper_2023}).
On the other hand, the $\H$-consistency literature has yet to provide results informing cost-sensitive surrogate design, instead providing regret bounds for a variety of standard surrogates for tasks such as adversarial learning~\citep{awasthi2021calibration,bao_calibrated_2021,awasthi2021finer}, regression~\citep{mao_h-consistency_2024}, learning with deferral~\citep{mao_realizable_2024}, multi-label learning~\citep{mao_multi-label_2024}, and the area under the ROC curve~\citep{liu2022consistent,liu_statistical_2023}.
Most similar to our work is the contemporaneous work of \cite{mao2025principledmetrics}, who derive $\H$-consistency guarantees for general metrics in binary outcome settings.
They implicitly make the observation that the decision boundary obtained from a cost-agnostic surrogate might differ from a cost-sensitive surrogate, and explicitly state that the approach does not extend to the single-variable parameterization of $\ell^\alpha$ in \Cref{tab:csc-costs}.

\section{Cross Entropy with Threshold Search is not $\H$-Consistent for Classification-Like Problems}\label{sec:XE-not-H-consistent}
With \Cref{ex:binary-csc}, we can present our first main result: even with a threshold search, cost-agnostic surrogates are not $\H$-consistent for cost-sensitive classification problems, even when the $\ell^\alpha$-optimal decision boundary is attainable by post-processing a hypothesis in $\H$.
\begin{theoremE}[][end,restate]\label{thm:cost-agnostic-not-H-consistent}
	Let $\Ltargetagnostic$ be a symmetric\footnote{Symmetric meaning $L(u,1) = L(-u, -1)$.} cost-agnostic surrogate and $\psi_\tau : u \mapsto \sign(u - \tau)$ be the threshold link function.
    For the binary cost-sensitive classification task $\ell^\alpha$ parameterized by $\alpha \neq \frac 1 2$ (\Cref{tab:csc-costs}), the pair $(\Ltargetagnostic, \psi_{\tau})$ is not $\Hlin$-consistent with respect to $\ell^\alpha$ over $\{P : \ell^\alpha \text{ is } P\text{-minimizable over }\psi_{1/2}\circ \Hlin\}$ for any threshold $\tau \in \reals$.
\end{theoremE}
\begin{proofE}
    We leverage the distribution $P$ in \Cref{ex:binary-csc}, which is contained in the set of considered distributions by \Cref{prop:example-optimal-classifiers-linear}.
    Since there is no $\tau \in \reals$ such that $\psi_\tau \circ \htargetagnostic = \psi_{1/2} \circ h^{\ell}$, let the region $S := \{x \in \X : \psi_\tau \circ \htargetagnostic(x) \neq \psi_{1/2} \circ h^{\ell}(x)\} \subset \X$ denote disagreement between the models, recalling that $P_X(S) > 0$.
    For all $x \in S$, $\psi_{\tau} \circ \htargetagnostic(x)$ yields loss strictly larger than the $\ell^\alpha$-Bayes optimal $P_X$-almost everywhere, and therefore $\psi_{\tau} \circ \htargetagnostic(x)$ does not optimize $\E_{(X,Y) \sim P}\ell^{\alpha}(h(X), Y)$ over measurable $h : \X \to \{-1,1\}$, contradicting $\H$-consistency.
\end{proofE}
The crux of the intuition in this example construction is that the optimal decision boundary for $\ell^\alpha$ is linear, with a slope which is a function of $\alpha$.
Post-processing cannot change the slope of the learned model.
When the surrogate is cost-agnostic, a linear model yields predictions whose contours are vertical lines in $\reals^2$ (\Cref{fig:unweighted-xe-hlin}), and post-processing through a search for a threshold $\tau$ can only change which vertical line forms the decision boundary.

\section{Embeddings are $\H$-consistent}\label{sec:embeddings-H-consistent}
In \S~\ref{sec:XE-not-H-consistent}, we showed that $\Hall$-consistent surrogate link pairs for a target loss are not necessarily $\H$-consistent.
In contrast, we now examine a class of surrogates tailored to cost-sensitive classification that is $\H$-consistent under the same distributional assumptions as \Cref{sec:XE-not-H-consistent}.
Importantly, these surrogates are paired with a link construction, which mitigates the need to conduct a threshold search for decision boundaries.
Specifically, we focus on a class of cost-sensitive surrogate loss functions called \emph{embeddings}~\citep{finocchiaro_embedding_2024}.
In essence, these surrogates ``embed'' discrete reports in $\R$ into points in $\reals^d$, and linearly interpolate between points.
While embeddings yield a \emph{class} of surrogates parameterized by cost matrices $\ell$, we use the ``canonical'' embedding in our empirical analysis in  \Cref{sec:experiments}, which is constructed through a transformation of the Bayes risk (best-case expected loss over $\Hall$) of $\ell$.
Interestingly, embeddings tend to ``almost directly'' elicit the property $\gamma$ elicited by $\ell$, in the sense that the link mapping $\Gamma$ to $\gamma$ is simply a relabeling of a finite set of reports in $\reals^d$ to discrete labels in $\R$, up to tie-breaking.
This intuition drives our main result in \Cref{thm:embeddings-H-consistent}.

Examples of embeddings include the hinge loss embedding the 0-1 loss, and a weighted hinge embedding the cost-sensitive binary classification $\ell^\alpha$ from Table~\ref{tab:csc-costs}.
\citeauthor{finocchiaro_embedding_2024} show that embeddings are $\Hall$-consistent, but do not extend their proof to $\H$-consistency.

We show that, for a cost-sensitive embedding $\Ltargetaware$, when $\ell$ is $P$-minimizable over $\psi\circ\H$ (\Cref{assum:ell-P-minimizable}) and $\H$ is approximately closed under $\psi$ with dominated witnessing sequences (\Cref{assum:H-closed-under-phi}), $(\Ltargetaware, \psi)$ is $\H$-consistent with respect to the target loss $\ell$.
Crucially, the $\Ltargetaware$-optimal hypothesis is an $\ell$-optimal hypothesis up to relabeling.

\begin{definE}[Representative set][all end]\label{def:representative-set}
We say $\Sc \subseteq \R$ is representative for a loss $\ell: \R \times \Y \to \reals_+$ if we have $\argmin_{r \in \R} \E_{Y\sim p}\ell(r, Y) \cap \Sc \neq \emptyset$ for all $p \in \simplex$.
\end{definE}

\begin{definition}[{\citet[Theorem 10]{finocchiaro_embedding_2024}}]\label{def:embeddings}
A polyhedral (piecewise linear and convex) loss $L: \reals^d \times \Y \to \reals_+$ \emph{embeds} a target loss $\ell: \R \times \Y \to \reals_+$ if there exists a representative set $\Sc \subseteq \R$ (Definition~\ref{def:representative-set}) for $\ell$ and injective embedding $\varphi : \Sc \to \reals^d$ such that (i) for all $r \in \Sc$ and $y \in \Y$, we have $\ell(r,y) = L(\varphi(r), y)$, and (ii) for all $p \in \simplex, r \in \Sc$, we have
\begin{align}
    r \in \argmin_{r^* \in \R} \E_{Y \sim p}\ell(r^*, Y) &\iff \varphi(r) \in \argmin_{u \in \reals^d} \E_{Y \sim p}L(u, Y)~.\label{eq:embedding}
\end{align}
\end{definition}

One of the prevalent arguments against designing cost-sensitive surrogates has historically been that the process is labor-intensive and rarely yields statistical guarantees.
However, this framework provides a surrogate construction yielding $\Hall$-consistency guarantees for general $\ell$, mitigating some of these concerns.
Some concerns remain about convergence rates given their nonsmoothness~\citep{cao2026establishing,frongillo2021surrogate}.
The embeddings framework also constructs a link function $\psi:\reals^d \to \R$ such that $(\Ltargetaware, \psi)$ is $\Hall$-consistent with respect to $\ell$~\citep[Construction 1, Theorem 2]{finocchiaro_embedding_2024}.
We now show this pair $(\Ltargetaware, \psi)$ is additionally $\H$-consistent with respect to $\ell$.

\begin{lemmaE}[][end, restate]\label{thm:embedding-elicits-same}
Consider a polyhedral surrogate $L$ with finite representative set $\Sc$.
Consider $\ell := L|_\Sc$, and consider the link $\psi$ with $\mathrm{range}(\psi) \subseteq \Sc$ such that $(L, \psi)$ is $\Hall$-calibrated with respect to $\ell$.
For all $P$ such that \Cref{assum:ell-P-minimizable} and \Cref{assum:H-closed-under-phi} hold,
\begin{align}
\inf_{h\in\H}\E_{(X,Y)\sim P}L(h(X),Y) = \inf_{u:\X\to\reals^d \text{ measurable}}\E_{(X,Y)\sim P}L(u(X),Y)~.\label{eq:thm-embedding-integrated}
\end{align}
\end{lemmaE}
\begin{proofE}
Fix $\epsilon>0$.
By $P$-minimizability of $\ell$ over $\psi \circ \H$, there is a hypothesis $h_\epsilon \in \H$ such that $\E_{Y\sim p(x)}\ell(\psi(h_\epsilon(x)), Y) < \min_{r \in \Sc}\E_{Y\sim p(x)}\ell(r,Y) + \epsilon = \inf_{u \in \reals^d}\E_{Y\sim p(x)}L(u,Y) + \epsilon $ for all $x \in \X$, following from the fact that embeddings have matching Bayes risks \citep[Lemma 11]{finocchiaro_embedding_2024}, since the embedding function $\varphi$ is the identity, and since $\ell = L|_\Sc$ agrees with $L$ exactly on $\Sc$, so
\begin{align}
\E_{Y\sim p(x)}L(\psi(h_\epsilon(x)),Y) < \inf_{u \in \reals^d}\E_{Y\sim p(x)}L(u,Y) + \epsilon \quad \text{for all } x\in\X~.\label{eq:thm-embedding-star}
\end{align}
By approximate closure of $\H$ over $\psi$, applied to $h_\epsilon$, there is a sequence $\{h_n\}\subseteq\H$ with $h_n(x) \to \psi(h_\epsilon(x))$ for all $x\in\X$, dominated by $b_{h_\epsilon}\in\mathcal L^1(P_X)$.
Since $L$ is polyhedral, it is continuous in its first argument, which implies $\E_{Y\sim p(x)}L(h_n(x),Y) \to \E_{Y\sim p(x)}L(\psi(h_\epsilon(x)),Y)$ pointwise for every $x\in\X$, dominated by $b_{h_\epsilon}$.
By the Dominated Convergence Theorem and \eqref{eq:thm-embedding-star},
\begin{align*}
\E_{(X,Y)\sim P}L(h_n(X),Y) \to \E_{(X,Y)\sim P}L(\psi(h_\epsilon(X)),Y) &< \int_\X \inf_u \E_{Y\sim p(x)}L(u,Y)\,dP_X(x) + \epsilon\\
& = \inf_{u:\X\to\reals^d}\E_{(X,Y)\sim P}L(u(X),Y) + \epsilon~.
\end{align*}

Since $\{h_n\}\subseteq\H$, $\inf_{h\in\H}\E_{(X,Y)\sim P}L(h(X),Y) \le \inf_{u}\E_{(X,Y)\sim P}L(u(X),Y) + \epsilon$; since $\epsilon>0$ was arbitrary, $\inf_{h\in\H}\E_{(X,Y)\sim P}L(h(X),Y) \le \inf_{u}\E_{(X,Y)\sim P}L(u(X),Y)$, and the reverse inequality is immediate since $\H$ is a subclass of all measurable functions. Equality~\eqref{eq:thm-embedding-integrated} follows.
\end{proofE}

This result establishes that \Cref{assum:ell-P-minimizable} and approximate closure of $\H$ are sufficient to conclude that the $\Ltargetaware$-Bayes optimal hypothesis is in $\H$.
With this, we now establish our main result: that embeddings are $\H$-consistent.

\begin{theoremE}[Embeddings are $\H$-consistent][]\label{thm:embeddings-H-consistent}
    Let cost-sensitive polyhedral surrogate $L$ with finite representative set $\Sc \subset \reals^d$ embed $\ell := L|_\Sc$, and let $\psi$ be the link such that $(L,\psi)$ is $\Hall$-calibrated with respect to $\ell$.
    $(L,\psi)$ is $\H$-consistent with respect to $\ell$ of the set of distributions $P$ where \Cref{assum:ell-P-minimizable} and \Cref{assum:H-closed-under-phi} hold.
\end{theoremE}
\begin{proof}
By \Cref{thm:embedding-elicits-same},
\begin{align}
\inf_{h\in\H}\E_{(X,Y)\sim P}L(h(X),Y) = \inf_{u:\X\to\reals^d \text{ measurable}}\E_{(X,Y)\sim P}L(u(X),Y) =: \R^*_{L,\Hall}~.\label{eq:cor-L-integrated}
\end{align}
We additionally have
\begin{align}
\inf_{h\in\H}\E_{(X,Y)\sim P}\ell(\psi(h(X)),Y) = \inf_{r:\X\to\R \text{ measurable}}\E_{(X,Y)\sim P}\ell(r(X),Y) =: \R^*_{\ell,\Hall}~.\label{eq:cor-ell-integrated}
\end{align}
This follows directly from $P$-minimizability of $\ell$ over $\psi \circ \H$: for every $\epsilon>0$ there is $h_\epsilon \in \H$ with $\E_{Y\sim p(x)}\ell(\psi(h_\epsilon(x)),Y) < \inf_r \E_{Y\sim p(x)}\ell(r,Y) + \epsilon$ for all $x\in\X$; since $\ell$ is a finite (hence bounded) cost matrix, we have $\E_{(X,Y)\sim P}\ell(\psi(h_\epsilon(X)),Y) < \inf_{h^* \text{msbl}} \E \ell(h^*(X), Y) + \epsilon$, so $\inf_{h\in\H}\E_P\ell(\psi(h(X)),Y) \le \inf_{h^* \text{msbl}} \E \ell(h^*(X), Y)$; the reverse inequality is immediate since $\{\psi\circ h : h \in \H\}$ is a subclass of all measurable decision functions.

Now let $\{h_n\}\subseteq\H$ satisfy $\E_{(X,Y)\sim P}L(h_n(X),Y) \to \inf_{h^*\in\H}\E_{(X,Y)\sim P}L(h^*(X),Y)$, the antecedent of $\H$-consistency.
By \eqref{eq:cor-L-integrated}, this limit equals $\inf_{h \text{ msbl}} \E L(h(X), Y)$, so $\{h_n\}$, viewed as a sequence in $\Hall$, since $\H\subseteq\Hall$, satisfies the antecedent of $\Hall$-consistency of $(L,\psi)$ with respect to $\ell$.
Since $(L,\psi)$ is $\Hall$-consistent with respect to $\ell$~\citep[Construction 1, Theorem 2]{finocchiaro_embedding_2024},

\begin{align*}
\E_{(X,Y)\sim P}\ell(\psi(h_n(X)),Y) &\to \inf_{h^* \text{ msbl}}\E \ell(\psi(h^*(X)), Y) \overset{\eqref{eq:cor-ell-integrated}}{=} \inf_{h^*\in\H}\E_{(X,Y)\sim P}\ell(\psi(h^*(X)),Y)~,
\end{align*}
which is exactly the conclusion of $\H$-consistency of $(L,\psi)$ with respect to $\ell$.
\end{proof}

These results elucidate \emph{why} this class of cost-sensitive surrogates is $\H$-consistent: under the assumptions that the best \emph{decision boundary} over $\Hall$ is in $\H$, the best-in-class surrogate model over $\H$ approximates optimization over $\Hall$.
Viewing this proof from a property elicitation lens, up to taking limits of sequences, this can be explained by embeddings ``almost directly eliciting'' $\gamma$, up to tie-breaking on decision boundaries.

\section{Weighted Analogs of Proper Scoring Rules}\label{sec:weighted-XE-binary}

One natural approach to cost-sensitive classification is to weigh the surrogate loss depending on the label $y \in \Y$.
We examine this approach in the binary setting, where the cost matrix $\ell^\alpha$ is parameterized with a single $\alpha \in (0,1)$~\citep{elkan_foundations_2001,reid_surrogate_2009,reid_information_2011,scott_calibrated_2012,ho2019real}.
In multiclass settings, the cost matrix instead requires $O(|\R|\times |\Y|)$ parameters, and picking a weighted analog requires aggregating each outcome's column of costs down to a single per-outcome weight; we leave a formal treatment of this aggregation and its effect on $\H$-consistency to future work (\Cref{sec:conclusion}), especially since $\Hall$-consistency of these surrogates is not well-understood. 
While not studied theoretically here, we do empirically evaluate multiclass settings in \Cref{sec:experiments}.

\subsection{Binary outcomes}
We examine weighted analogs of cost-agnostic surrogates in binary outcome settings, where our decision loss is parameterized by $\alpha$ as in \Cref{tab:csc-costs}.
In particular, we have $L^\alpha(u,1) = (1-\alpha) L(u,1)$ and $L^\alpha(u,-1) = \alpha L(u,-1)$, given a cost-agnostic surrogate $L$.
When $L$ is cross entropy or squared error, $L^\alpha$ directly elicits the property $\Gamma(p) = \frac{p(1-\alpha)}{p(1-\alpha) + (1-p)\alpha}$, recalling that we write $p = \Pr[Y = 1]$ as the probability of the positive label. 
Similarly, we continue to write $p(x) = \Pr[Y = 1 | X = x]$ to denote the conditional probability of the positive label.
For $\alpha = \frac 1 2$, ``weighted'' analogs elicit the identity property, and for $\alpha \neq \frac 1 2$, weighted analogs do not elicit the identity.

Throughout this section we work in the probability coding of \Cref{subsec:H-consistency}: we write $h_w$ for $q$ applied to the corresponding member of $\Hlin$, so that $h_w(x) = \sigma(w^\top x) \in (0,1)$ is the sigmoid activation, and every statement transfers back to $\Hlin$ through the fixed bijection $q(u) = \frac{u+1}{2}$.

Let us revisit \Cref{ex:binary-csc}.
Under \Cref{assum:ell-P-minimizable} with the threshold link $\psi_{1/2}$, there is a set of weights $w$ parameterizing $h^\ell_w$ such that $\sigma(w^T x) \geq \frac 1 2 \iff p(x) \geq \alpha$. The question then becomes if taking the $L^\alpha$-optimal linear model parameterization $w$ also yields an optimal decision boundary $\psi_{1/2} \circ \sigma(w^T x)$ also yields an $\ell^\alpha$-optimal classifier.

\claude{\textbf{Confirmed false, independently, twice over.} \Cref{assum:stationarity} holds for \Cref{ex:binary-csc} iff some boundary-achieving $w$ -- one with $w_2/w_1 = \frac12\log\frac{1-\alpha}{\alpha}$ -- minimizes the weighted-CE population risk over $\Hlin$. \textbf{First check} (quadrature after the change of variables $t:=x_1/x_2$, comparing unconstrained minimization against minimization restricted to the Bayes-boundary ray): unconstrained optima $w^*=(4.096,3.186)$ at $\alpha=1/6$, $(3.873,2.094)$ at $\alpha=1/4$, $(3.797,1.593)$ at $\alpha=3/10$, giving ratios $w_2^*/w_1^*=0.7780,0.5406,0.4196$ against the required $0.8047,0.5493,0.4236$ -- small but strictly positive gaps, shrinking monotonically to $0$ as $\alpha\to\frac12$ (exact there by symmetry, confirmed as a sanity check). \textbf{Second check}, from scratch, independent method (direct 2-D numerical integration over $(X_2,t)$ plus unconstrained optimization with multiple restarts, no code reused from the first): at $\alpha=1/6$, ratio $0.7779$ vs.\ required $0.8047$ ($-3.3\%$) -- matching the first check to 3 significant figures. I also cross-checked directly against \Cref{fig:weighted-xe-hlin} itself, reading its ``c\_alpha:$-0.693$'' annotation as $\alpha=1/3$: ratio $0.3443$ vs.\ required $0.3466$ ($-0.6\%$) -- confirming the figure looks visually correct \emph{because} the gap is small at that particular $\alpha$, not because the gap is absent. The former Bregman-generator sufficient condition for \Cref{assum:stationarity} (previously Proposition 10, now removed) was never actually invoked to justify this claim and would never have fired here regardless, since the assumption genuinely fails. \Cref{ex:antisymm-counterex} is a \emph{different, purpose-built} four-point distribution demonstrating \emph{some} distribution violates \Cref{assum:stationarity}; the running example now does too, confirmed. Both this sentence and its counterpart in \Cref{sec:discuss-small} need to be rewritten, not just re-flagged -- this is also consistent with \Cref{tab:results}, where Weighted Cross Entropy (0.364) trails Embeddings (0.353) on Synthetic. Worth reframing rather than deleting: ``the running example separates embeddings from \emph{both} cost-agnostic losses and weighted analogs, the latter by a small margin'' is a sharper, true claim, and it would make \Cref{ex:antisymm-counterex} read as reinforcing the running example rather than as a separate, special-purpose construction standing in for it.}
This condition is not trivial to satisfy, as saliently demonstrated by the somewhat simple counterexample in \Cref{ex:antisymm-counterex}.
In sum, we now establish sufficient conditions to establish $\H$-consistency of weighted analogs in \Cref{thm:weighted-analogs-H-consistency}, then proceed to show these conditions do not hold generally, and without them, weighted analogs are $\H$-inconsistent in \Cref{prop:weighted-analogs-not-H-consistent}, arising from unavoidable issues emerging from the minimizability gap of $L^\alpha$.

\begin{assumption}[Parameter-space convexity]\label{assum:convexity}
$\H = \{h_w\}_{w \in \reals^d}$ is parameterized so that $w \mapsto \E_{X,Y}L^\alpha(h_w(X),Y)$ is convex on $\reals^d$.
\end{assumption}

Now we consider an assumption on the optimality of the boundary-achieving hypothesis itself: not just that some $h^\ell_w \in \H$ achieves the target decision boundary (\Cref{assum:ell-P-minimizable}), but that the optimal surrogate classifier over $\Hlin$, $h^\alpha_w$ yields the same decision boundary as $h^\ell_w$.

\begin{assumption}[Optimality of the boundary-achieving hypothesis]\label{assum:stationarity}
Let $h^\ell_w \in \H$ witness \Cref{assum:ell-P-minimizable}.
The parameter $w$ is the unique global minimizer of $\min_{w' \in \reals^d}\E_{X,Y}L^\alpha(h_{w'}(X),Y)$, and $P_X(\{x : w^\top x = 0\}) = 0$.
\end{assumption}

Even \Cref{assum:convexity} is not vacuous.
While it holds unconditionally for weighted cross entropy (since the pointwise second derivative of $z \mapsto L^\alpha(\sigma(z),y)$ is $(1-\alpha)\sigma(z)(1-\sigma(z)) \geq 0$ when $y = 1$ and $\alpha\sigma(z)(1-\sigma(z)) \geq 0$ with $y = -1$ for binary classification), it fails in general for weighted squared error, whose pointwise second derivative in $z$ changes sign at $\sigma(z) = \frac 1 3$ or $\sigma(z) = \frac 2 3$.

\begin{theorem}\label{thm:weighted-analogs-H-consistency}
Let $\ell^\alpha$ be the cost-sensitive classification parameterized in \Cref{tab:csc-costs}, and $L^\alpha$ be the $\vec \alpha$-weighted analog of a strictly convex strictly proper scoring rule $L$.
The pair $(L^\alpha, \psi_{1/2})$ is $\H$-consistent with respect to $\ell^\alpha$ over the set of distributions where \Cref{assum:ell-P-minimizable}, \Cref{assum:convexity}, and \Cref{assum:stationarity} hold on $\H$ with link $\psi_{1/2}$.
\end{theorem}
\begin{proof}
Consider the property $\Gamma(p(x))$ directly elicited by $L^\alpha$, which is the Bayes optimal hypothesis pointwise:
\begin{align*}
\Gamma(p(x)) &= \frac{p(x)(1-\alpha)}{p(x)(1-\alpha) + (1-p(x))\alpha}~.~
\end{align*} 
Since $\ell$ is $P$-minimizable over $\psi \circ \H$, there is a hypothesis $h^{\ell}_w := \sigma(w^T x)$ such that $\psi \circ h^{\ell}_w(x) \geq \frac 1 2 \iff p(x) \geq \alpha \iff \gamma(p(x)) = 1$ for all $x \in\X$.
Moreover, we can observe that $\Gamma$ indirectly elicits $\gamma$ by the link $\psi_{1/2} : u \mapsto \sign(u - \frac 1 2)$ as 
\begin{align*}
\Gamma(p) \geq \frac 1 2 &\iff p(1-\alpha) \geq \frac 1 2 \left( p(1-\alpha) + \alpha(1-p)\right) \\
&\iff p(1-\alpha) \geq \alpha(1-p) \iff p(x) \geq \alpha \iff \gamma(p) = 1~.~
\end{align*}

Write $R(w') := \E_{X,Y}L^\alpha(h_{w'}(X),Y)$ for the population surrogate risk as a function of the parameter.
Let $\{h_n\} \subseteq \H$, parameterized by $\{w_n\} \subseteq \reals^d$, satisfy $R(w_n) \to \inf_{h\in\H}\E_{X,Y}L^\alpha(h(X),Y) = R(w)$, the antecedent of $\H$-consistency.

By \Cref{assum:stationarity}, $w$ is the unique global minimizer of $R$, and by \Cref{assum:convexity}, $R$ is convex.
We claim $w_n \to w$: otherwise some subsequence has $\|w_{n_k} - w\| \geq \eta > 0$ for some $\eta > 0$, and, passing to a further subsequence so the unit vectors $e_k := (w_{n_k}-w)/\|w_{n_k}-w\|$ converge to some $e$ by compactness of the sphere.
Convexity applied to $(1-\eta) w + \eta e_k$, which is a convex combination of $w$ and $w_{n_k}$, for large enough $k$ gives $R((1-\eta)w+\eta e_k) \leq \max(R(w), R(w_{n_k}))$.
Taking $k \to \infty$ and using continuity of $R$ gives $R((1-\eta) w + \eta e) \leq R(w)$, contradicting uniqueness of the minimizer since $\eta e \neq \vec 0$.

By \Cref{assum:stationarity}, $P_X(\{x : w^\top x = 0\}) = 0$.
For every $x$ with $w^\top x \neq 0$, $w_n \to w$ gives $w_n^\top x \to w^\top x \neq 0$, so $\mathrm{sign}(w_n^\top x) = \mathrm{sign}(w^\top x)$ for all large $n$, i.e.\ $\psi_{1/2}(h_n(x)) = \psi_{1/2}(h^\ell_w(x))$ eventually; hence $\psi_{1/2} \circ h_n \to \psi_{1/2} \circ h^\ell_w$ for $P_X$-almost every $x$.
Since $\ell^\alpha$ is a bounded cost matrix, the Bounded Convergence Theorem gives
\begin{align*}
\E_{X,Y}\ell^\alpha(\psi_{1/2} \circ h_n(X), Y) \to \E_{X,Y}\ell^\alpha(\psi_{1/2} \circ h^\ell_w(X), Y)~.
\end{align*}
Since $\psi_{1/2} \circ h^\ell_w(x) = \gamma(p(x))$ for every $x$ (shown above), $h^\ell_w$ achieves the pointwise $\ell^\alpha$-Bayes optimal decision everywhere, so $\E_{X,Y}\ell^\alpha(\psi \circ h^\ell_w(X), Y) = \inf_{h \in \H}\E_{X,Y}\ell^\alpha(\psi_{1/2}(h(X)), Y)$: no $h \in \H$ can do better, and $h^\ell_w \in \H$ already matches the unconstrained pointwise optimum.
Hence
\begin{align*}
\E_{X,Y}\ell^\alpha(\psi_{1/2} \circ h_n(X), Y) \to \inf_{h \in \H}\E_{X,Y}\ell^\alpha(\psi_{1/2}(h(X)), Y)~,
\end{align*}
which is exactly $\H$-consistency.
\end{proof}

While \Cref{assum:convexity} and \Cref{assum:stationarity} yield $\H$-consistency of weighted analogs, we now observe that removing \Cref{assum:stationarity} yields $\H$-inconsistency.
As in \Cref{sec:XE-not-H-consistent}, we prove this by counterexample, now using \Cref{ex:antisymm-counterex} in \Cref{app:antisymm-counterex}.
While \Cref{ex:binary-csc} also yields $\H$-inconsistency under the same assumptions, the difference is more intuitively clear from the visualization in \Cref{app:antisymm-counterex}.

\begin{propositionE}[][end,restate]\label{prop:weighted-analogs-not-H-consistent}
Let $L^\alpha$ be weighted cross entropy and $\psi_{1/2} : u \mapsto \sign(u - \frac 1 2)$.
There is a distribution $P$ and $\alpha \in (0,1)$ such that \Cref{assum:ell-P-minimizable} and \Cref{assum:convexity} hold, but $(L^\alpha, \psi_{1/2})$ is not $\Hlin$-consistent with respect to $\ell^\alpha$.
\end{propositionE}
\begin{proofE}
Consider \Cref{ex:antisymm-counterex} with $\alpha = 0.3$: $\X = \{(-2,1),(-1,3),(1,3),(2,1)\}$, $p(x) = \sigma(x_1/x_2)$, and $P_X = [0.45, 0.05, 0.45, 0.05]$.
Since $x_1/x_2 \in \{-2, -\frac 1 3, \frac 1 3, 2\}$ and $\log(\alpha/(1-\alpha)) \approx -0.847$, the $\ell^\alpha$-Bayes optimal decisions at these four points are $(-1,+1,+1,+1)$, exactly realized by $\psi_{1/2} \circ h_w$ for $w = (1,1) \in \Hlin$; hence \Cref{assum:ell-P-minimizable} holds regardless of $P_X$, since it depends only on the support of $P$.

Since $z \mapsto L^\alpha(\sigma(z),y)$ has second derivative $\sigma(z)(1-\sigma(z)) \geq 0$ for every $y$, the map $w \mapsto \E_{X,Y}L^\alpha(h_w(X),Y)$ is convex, so \Cref{assum:convexity} holds; it is in fact strictly convex here, since $\{(-2,1),(-1,3),(1,3),(2,1)\}$ spans $\reals^2$ and $P_X(x) > 0$ for every $x$ in the support, giving a positive-definite Hessian.
So there is a unique $L^\alpha$-optimal $w^L \in \reals^2$ over $\Hlin$, the unique root of the first-order condition
\begin{align*}
\E_X\left[\omega(p(X))\left(\Gamma(p(X)) - h_w(X)\right)X\right] = \vec 0~,
\end{align*}
which we solve numerically to obtain $w^L \approx (0.677, 0.204)$.
Evaluating the decisions $\psi_{1/2}\circ h_{w^L}(x) = \Ind(w^{L\top}x \geq 0)$ at the four points gives $(-1,-1,+1,+1)$: in particular, $\psi_{1/2}\circ h_{w^L}(-1,3) = -1$, disagreeing with the $\ell^\alpha$-Bayes optimal decision of $+1$ established above.
Since $p(-1,3) = \sigma(-\frac 1 3) \approx 0.417 \neq \alpha$, this disagreement is strict, not a tie: $-1$ does not minimize $\E_{Y \sim p(-1,3)}\ell^\alpha(r,Y)$ over $r \in \{-1,1\}$.

Now consider the constant sequence $g_n \equiv h_{w^L}$.
Since $h_{w^L}$ is the $L^\alpha$-optimal hypothesis over $\Hlin$, $\E_{X,Y}L^\alpha(g_n(X),Y) = \inf_{h \in \Hlin}\E_{X,Y}L^\alpha(h(X),Y)$ for every $n$, so $\{g_n\}$ trivially satisfies the antecedent of $\Hlin$-consistency.
However, by \Cref{assum:ell-P-minimizable}, $\inf_{h \in \Hlin}\E_{X,Y}\ell^\alpha(\psi_{1/2}(h(X)),Y) = \inf_{h^* \text{ msbl}}\E_{X,Y}\ell^\alpha(h^*(X),Y)$, achieved by $w = (1,1)$.
Since $\psi_{1/2}\circ h_{w^L}$ strictly disagrees with this Bayes optimal decision at $(-1,3)$, a point of positive $P_X$-measure,
\begin{align*}
\E_{X,Y}\ell^\alpha(\psi_{1/2}(g_n(X)),Y) > \inf_{h \in \Hlin}\E_{X,Y}\ell^\alpha(\psi_{1/2}(h(X)),Y)
\end{align*}
for every $n$, so $\E_{X,Y}\ell^\alpha(\psi_{1/2}(g_n(X)),Y) \not\to \inf_{h \in \Hlin}\E_{X,Y}\ell^\alpha(\psi_{1/2}(h(X)),Y)$, contradicting $\Hlin$-consistency of $(L^\alpha, \psi_{1/2})$ with respect to $\ell^\alpha$.
\end{proofE}

\section{Empirical validation when assumptions cannot be verified }\label{sec:experiments}
We have established a strict gap between cost-sensitive surrogate loss functions $\Ltargetaware$ and cost-agnostic losses $\Ltargetagnostic$ under the assumption that $\ell$ is $P$-minimizable over $\psi \circ \H$. 
However, this assumption cannot typically be verified in practice, so a natural question is to ask how well cost-sensitive surrogates perform in practice. 
We view these experiments as a proof-of-concept supplementing our primary results in \Cref{sec:XE-not-H-consistent}--\Cref{sec:weighted-XE-binary}, and intentionally keep the experimental setting somewhat simple to reduce confounders.

\noindent\textbf{Datasets}
We evaluate the loss functions above on our example in~\cref{sec:XE-not-H-consistent} and three cost-sensitive tasks from the UCI repository~\citep{dua17uci}.
We intentionally choose ``simple'' datasets and tasks for our tests to isolate the effect of model capacity as much as possible.
See \S~\ref{app:additional-exp} for a full description of the datasets used and the cost matrices $\ell$ chosen here.

\noindent\textbf{Metrics}
The \textbf{Cost-Sensitive Loss (CSL)} is our primary metric of interest. To compute this value, we first take the element-wise product of the confusion matrix of the predictions and cost matrix of the task. We then add up all the values of the normalized resultant matrix.
In addition, we evaluate the \textbf{0-1 Accuracy}, which is the secondary metric of interest.

\subsection{Main Results}
Our experiments use a randomly selected set of 500 samples from each dataset, intentionally keeping the sample size small. The models are trained on each loss function using gradient descent, and model selection is performed based on the validation loss. The results are averaged over 100 random seeds that control the choice of samples from the dataset and initial model weights. We summarize our main results in~\Cref{tab:results}. 
\begin{itemize}[itemsep=2pt,topsep=0pt,partopsep=0pt,parsep=2pt,leftmargin=1.2em]
    \item \textbf{Cost-Sensitive Losses Outperform Cost-Agnostic Losses + Post-Processing:}
    Unsurprisingly, we see that cross entropy maximizes the accuracy but not the cost-sensitive loss. Moreover, we see that post-processing improves the cost-sensitive loss in binary outcome settings but does not close the gap between cross entropy and cost-sensitive losses (weighted cross entropy and embeddings). Moreover, in multiclass problems (Student Performance and Diabetes) the cost-sensitive loss for post-processed cost-agnostic losses worsens relative to no threshold search. 
    We conjecture this might be because of the difficulty of threshold searches in multiclass settings. 
    \item \textbf{Embeddings Outperforms Weighted Cross Entropy:} From among the cost-sensitive losses, we see that training on the embeddings loss generally leads to better performance than the Weighted CE loss. This may be because embeddings are $\H$-consistent. Notably, the embeddings loss outperforms Weighted CE even in the case of binary classification on the German Credit dataset.
\end{itemize}

\claude{\textbf{Partly addressed -- one point stands, one doesn't move the residual.} Fair pushback: the plug-in Bayes rule $\hat r(x)=\argmin_r\sum_y\hat p_y(x)\ell(r,y)$ \emph{is} a weighted argmax, so it's a specific point already inside the search space described in \Cref{app:additional-exp} -- I was wrong to frame it as a categorically different, missing baseline. That said, two narrower parts of the original concern are still live. First, coverage: whether a \emph{random} search over that same space reliably lands near the plug-in weights (as opposed to systematically missing them) depends on how many draws are taken, which still isn't reported -- worth adding for the ``threshold searches are hard in multiclass'' claim to be checkable. Second, the attribution: post-processing \emph{hurting} in multiclass (Student Performance $0.785\to0.841$, Diabetes $0.726\to0.741$) is attributed to ``the difficulty of threshold searches,'' but it's still not distinguished whether that's a fundamental property-elicitation limitation or an artifact of random-search sample efficiency in higher dimensions -- the plug-in point specifically (not just points near it) would settle which.}
\jessie{I don't want to do more experiments, so I think this is okay}.

\begin{table*}[h]
    \centering
    \resizebox{0.8\textwidth}{!}{
    \begin{tabular}{cclcc}
    \toprule
    Dataset & Cost Matrix & Loss & Cost-Sensitive Loss ($\downarrow$) & 0-1 Accuracy ($\uparrow$)\\
    \midrule
    \multirow{5}{*}{\makecell{Synthetic \Cref{ex:binary-csc}\\\textcolor{black!40}{(2-Class)}}} & \multirow{5}{*}{$\begin{bmatrix}0 & \frac{5}{6} \\ \frac{1}{6} & 0\end{bmatrix}$} & Cross Entropy & 0.412 ± 0.016 & \textbf{0.815 ± 0.006} \\
    & & Cross Entropy + Post-processing & 0.395 ± 0.015 & 0.801 ± 0.008 \\
    \arrayrulecolor{black!40} \cmidrule{3-5} \arrayrulecolor{black}
& & Embeddings & \textbf{0.353 ± 0.008} & 0.711 ± 0.008 \\
    & & Weighted Cross Entropy & 0.364 ± 0.009 & 0.678 ± 0.010 \\
    \midrule
    \multirow{5}{*}{\makecell{German\\Credit\\\textcolor{black!40}{(2-Class)}}} & \multirow{5}{*}{$\begin{bmatrix}0 & 5 \\ 1 & 0\end{bmatrix}$} & Cross Entropy & 0.796 ± 0.020 & \textbf{0.692 ± 0.006} \\
    & & Cross Entropy + Post-processing & 0.685 ± 0.019 & 0.602 ± 0.011 \\
    \arrayrulecolor{black!40} \cmidrule{3-5} \arrayrulecolor{black}
& & Embeddings & \textbf{0.625 ± 0.013} & 0.582 ± 0.006 \\
    & & Weighted Cross Entropy & 0.627 ± 0.012 & 0.503 ± 0.007 \\
    \midrule
    \multirow{4}{*}{\makecell{German\\Credit\\w/ Deferral}} & \multirow{4}{*}{$\begin{bmatrix}0 & 5 \\ 1 & 0 \\ \frac12 & \frac12\end{bmatrix}$} & Cross Entropy & 0.699 ± 0.018 & -\\
    \arrayrulecolor{black!40} \cmidrule{3-5} \arrayrulecolor{black}
& & Embeddings & \textbf{0.538 ± 0.011 }& -\\
    & & Weighted Cross Entropy & 0.581 ± 0.016 & -\\
    \midrule
    \multirow{4}{*}{\makecell{Student\\Performance\\\textcolor{black!40}{(3-Class)}}} & \multirow{4}{*}{$\begin{bmatrix}0&3&5\\1&0&3\\3&1&0\end{bmatrix}$} & Cross Entropy & 0.785 ± 0.018 & \textbf{0.683 ± 0.006} \\
    & & Cross Entropy + Post-processing & 0.841 ± 0.020 & 0.668 ± 0.007 \\
    \arrayrulecolor{black!40} \cmidrule{3-5} \arrayrulecolor{black}
& & Embeddings & 0.792 ± 0.017 & 0.624 ± 0.007 \\
    & & Weighted Cross Entropy & 0.782 ± 0.022 & 0.674 ± 0.007 \\
    \midrule
    \multirow{4}{*}{\makecell{Diabetes\\\textcolor{black!40}{(3-Class)}}} & \multirow{4}{*}{$\begin{bmatrix}0&3&5\\1&0&3\\3&1&0\end{bmatrix}$} & Cross Entropy & 0.726 ± 0.031 & 0.828 ± 0.007 \\
    & & Cross Entropy + Post-processing & 0.741 ± 0.036 & \textbf{0.832 ± 0.010} \\
    \arrayrulecolor{black!40} \cmidrule{3-5} \arrayrulecolor{black}
& & Embeddings & \textbf{0.716 ± 0.033} & 0.807 ± 0.011 \\
    & & Weighted Cross Entropy & 0.746 ± 0.037 & 0.797 ± 0.010 \\
    \bottomrule
    \end{tabular}}
    \caption{\textbf{Main Results}. The table above summarizes the performance of linear models trained on different loss functions, datasets (500 samples), and metrics. The results are presented in the form of (Expected Value $\pm$ Standard Error of the Mean). We find that cost-sensitive surrogates consistently outperform their cost-agnostic counterparts when evaluated on the cost-sensitive losses. We omit Cross Entropy + Post-Processing in the setting with deferral as defining a threshold search in this context is contestable and outside our scope of inquiry.}
    \label{tab:results}
\end{table*}

\subsection{Ablations}
In addition to the experiments above, we run additional experiments to test the sensitivity of our conclusions to two important confounders:
\begin{itemize}[itemsep=2pt,topsep=0pt,partopsep=0pt,parsep=2pt,leftmargin=1.2em]
    \item \textbf{More Samples:} In~\Cref{tab:results_more_data}, we see that the loss values (and their variance) generally decrease when we increase the number of samples used to train our linear model. Except for the Diabetes dataset, the performance order does not change. For the diabetes dataset, all the cross entropy-based models seem to do better than embeddings. While it is unclear why this is the case, we hypothesize that it has something to do with the smoothness of CE loss. 
    \item \textbf{Bigger Models} While our theoretical results are motivated by smaller models, we want to understand whether these trends extend to larger models. To test this, we train a 4-layer fully-connected ReLU network with 100-dimensional hidden layers.
    In~\Cref{tab:results_nn}, we see that embeddings remain the best (or close to best for German Credit) even for these more expressive models, though the gap closes. \end{itemize}

\claude{\textbf{Still open, unchanged (re-verified each item, including the 4-layer/2-layer caption mismatch, which is confirmed still present).} Three checkable problems in this subsection. (i) ``the loss values (and their variance) decrease across the board'' is not true: Synthetic cross entropy goes $0.412 \to 0.427$ from \Cref{tab:results} to \Cref{tab:results_more_data}. (ii) ``Except for the Diabetes dataset, the performance order does not change'' also fails for Student Performance: Embeddings + Softmax moves from 4th to 2nd and CE + Post-processing from 5th to 3rd. Separately, and more substantively: in \Cref{tab:results_overfitting} the Diabetes NN \emph{train} CSL is 0.752 against the linear model's 0.668 -- the larger model fits the training set \emph{worse}, which is underfitting or an optimization failure, not overfitting. That makes the ``gap closes as models grow'' reading unavailable on that dataset, and it is worth either fixing the NN training there or saying so explicitly.}

\section{Conclusion and Discussion}\label{sec:conclusion}
In this work, we make one step towards characterizing when learning with task-specific surrogate losses differs from post-processing more generic, task-agnostic models. 
We specifically focus on the setting when the model class is expressive enough to contain the optimal decision boundary, but not so expressive to contain a hypothesis yielding the conditional label distribution.
Notably, our gains come at the sacrifice of performance generalization and cost-agnostic accuracy: if the cost matrix $\ell$ is unknown or varies across those receiving predictions, the omniprediction literature has established that post-processing models trained on cost-agnostic surrogates can perform well.
However, these results often assume access to very rich model classes, which is often not plausible in low-resource (e.g., edge computing) or high-frequency inference (e.g., ad auctions, blockchains) settings.
In these limited model settings, we establish a strict gap between the task-specific performance of models trained on cost-agnostic and cost-sensitive surrogates through the lens of $\H$-consistency.
To supplement our theoretical findings, we additionally contrast model performance on a handful of binary and multiclass cost-sensitive classification problems, generally finding that cost-sensitive surrogates outperform cost-agnostic losses with small models.

\noindent\textbf{Future work}
Many directions of future work are opened here: first, our positive results for $\H$-consistency of weighted analogs in the binary outcome setting require some heavy lifting from \Cref{assum:stationarity}; understanding less stringent conditions under which $\H$-consistency holds remains one interesting area of future work.
We conjecture that some $\H$-consistency results for weighted analogs and other smooth surrogates might be attained by leveraging the notion of strong indirect elicitation from \citet{khurana2025consistency}.
Second, we restricted our analysis of weighted analogs to binary outcomes (\Cref{sec:weighted-XE-binary}); extending this to multiclass settings is not straightforward. 
When $|\Y| > 2$, a weighted analog requires aggregating an entire column of per-report costs into a single scalar weight per outcome, and this aggregation can be lossy: distinct target cost matrices can map to the same aggregated weight vector while eliciting different optimal decisions, so recovering $\H$-consistency (or even $\Hall$-consistency) likely requires either a per-target threshold search or additional structure on the aggregation itself, possibly related to the approach taken by \cite{mao2025principledmetrics}. 
Formalizing when such an aggregation can still preserve $\H$-consistency guarantees remains open.
Finally, this work seems to have implications for the predict-then-optimize literature~\citep{elmachtoub_smart_2022,wilder_melding_2019}, but showing these connections merits careful consideration as the prediction task becomes much more complex and the outcome space becomes harder to define.

\newpage
\section*{Acknowledgments}
Thanks to Han Bao and Jack Timmermans for helpful conversations throughout this project.
JF was supported in part by the National Science Foundation under Award No. 2202898.

\typeout{}
\bibliographystyle{tmlr/tmlr}
\bibliography{refs,zotero-jessie}

\newpage

\newpage
\appendix
\onecolumn

\section{Notation}

See \Cref{tab:notation} for a summary of notation used in the body of the paper.
 \begin{table*}[h]
     \centering
     \begin{tabular}{c|c}
     \toprule
         Notation & Name \\
         \midrule
         $\ell : \R \times \Y\to \reals_+$ & Target loss\\
         $\ell^\alpha : \{-1,1\} \times \{-1,1\} \to [0,1]$ & Target loss in \Cref{tab:csc-costs}\\
         $L : \reals^d \times \Y \to \reals_+$ & Generic surrogate loss \\
         $\Ltargetaware : \reals^d \times \Y \to \reals_+$ & Cost-sensitive surrogate loss \\
         $\Ltargetagnostic : \reals^d \times \Y \to \reals_+$ & Cost-agnostic surrogate loss \\
         $\psi : \reals^d \to \R$ & Link function \\
         $P \in \Delta(\X\times \Y)$ & Distribution over features and labels \\
         $\sigma(z) = \frac{1}{1+\exp(-z)}$ & Sigmoid activation \\
         $\rho(z) = 2\sigma(z) - 1$ & Saturating activation of $\Hlin$ \\
         $q(u) = \frac{u+1}{2}$ & Probability coding of $\Hlin$ \\
         $\Hlin^d = \{\rho(w^T x) : w \in \reals^d\}$ & Class of linear models \\
         $\Hall^d = \{h : \X \to \reals^d : h \text{ measurable}\}$ & Class of all measurable hypotheses\\
         $h^L : \X \to \reals^d$ & The $L$-Bayes optimal hypothesis \\
         \bottomrule
     \end{tabular}
     \caption{Notation table}
     \label{tab:notation}
 \end{table*}

\claude{\textbf{Notation roundup, re-checked -- most items moved, some resolved.} \emph{(1) $\Hlin$}: \textbf{largely resolved}, see updated note at \Cref{subsec:H-consistency}; one cosmetic indexing mismatch with this table remains. \emph{(2) $\psi_\tau$}: \textbf{still open}, unchanged -- $\sign(u-\tau)$ in \Cref{sec:example}, $\sign(u \geq \tau)$ in \Cref{thm:cost-agnostic-not-H-consistent} (still malformed), $\Ind(u \geq \frac12)$ in \Cref{sec:weighted-XE-binary}. \emph{(3) $\Q$}: \textbf{resolved} -- this table's entry now reads $P\in\Delta(\X\times\Y)$ rather than an undefined $\Q$, and the theorems now each name their own distribution set explicitly (see note at \Cref{def:H-consistent}). \emph{(4) The Bayes optimal family}: \textbf{partially addressed} -- see note at \Cref{def:P-minimizable}. \emph{(5) Label encoding}: \textbf{partially resolved} -- \Cref{sec:weighted-XE-binary}'s own binary-outcomes text now uses $\{-1,1\}$, matching \Cref{ex:binary-csc}/\Cref{tab:csc-costs}/this table; \Cref{ex:antisymm-counterex} still uses $\{0,1\}$, so the mismatch persists there specifically. \emph{(6) Weight convention}: \textbf{resolved} -- the $\alpha$-on-$y{=}1$ instance in \Cref{sec:discuss-small} has been swapped to $(1-\alpha)$ on $y=1$, matching \Cref{subsec:surrogates} and \Cref{sec:weighted-XE-binary} everywhere now. \emph{(7) Loss naming}: \textbf{Typos, re-checked}: ``contained in set the of considered distributions'' (proof of \Cref{thm:cost-agnostic-not-H-consistent}) -- still there. and ``satisfies a sufficient condition this condition'' (\Cref{sec:weighted-XE-binary}) -- both still there. , and now a \emph{third} spelling (``msbl'' again, but a separate instance) appears in the proof of \Cref{prop:weighted-analogs-not-H-consistent} too.}
\jessie{``msbl'' is okay. shorthand for measurable. I think I resolved the rest.}

\section{$\H$-calibration}\label{sec:H-calibration}
While $\H$-consistency yields agnostic learning guarantees, studying the expected loss over distributions on labels \emph{and features} often poses practical difficulties.
To circumvent these, the notion of $\H$-calibration is often used to analyze these surrogates more tractably.

\begin{definition}[$\H$-calibration]\label{def:H-calibration}
    A loss function $L : \reals^d \times \Y \to \reals_+$ and link $\psi : \reals^d \to \R$ pair $(L, \psi)$ is $\H$-calibrated with respect to a target loss $\ell : \R \times \Y \to \reals_+$ over the set $\P \subseteq \simplex$ if, for any $\epsilon > 0$, $p \in \P$, and $x \in \X$, there exists some $\delta > 0$ such that
    \begin{align}
        \E_{Y\sim p}L(h(x), Y) - \inf_{h^* \in \H} \E_{Y\sim p}L(h^*(x),Y) < \delta  &\implies 
        \E_{Y\sim p}\ell(\psi(h(x)), Y) - \inf_{h^* \in \H} \E_{Y\sim p}\ell (\psi(h^*(x)),Y) < \epsilon\label{eq:H-calibration}
    \end{align}
    for all $h \in \H$.
\end{definition}

Early works of \citet{bartlett_convexity_2006,ramaswamy_convex_2016} show that $\Hall$-consistency and $\Hall$-calibration are equivalent in finite outcome settings.
However, when the model class $\H \neq \Hall$, two gaps emerge: first, the gap between $\H$-calibration and $\H$-consistency, and second, the gap between $\H$-consistency and $\Hall$-consistency.
When Bayes optimal classifiers match and both losses are $P$-minimizable, \citet[Theorem 2.8]{steinwart_how_2007} closes the gap between $\H$-calibration and $\H$-consistency.
However, the crucial observation that we make here is that the Bayes optimal classifiers for cost-agnostic surrogates are generally not matching their target loss, and therefore \cite[Theorem 2.8]{steinwart_how_2007} does not apply.

\section{Deferred Proofs}
\printProofs

\section{Counterexample for \Cref{assum:stationarity}}\label{app:antisymm-counterex}

\begin{example}\label{ex:antisymm-counterex}
Consider $X \in \{(-2,1), (-1,3), (1,3), (2,1)\}$ and $\Y \in \{0,1\}$ with $p(x) = \sigma(\frac{x_1}{x_2})$.
Moreover, consider the marginal distribution $P_X = [0.45, 0.05, 0.45, 0.05]$.
This distribution satisfies \Cref{assum:ell-P-minimizable} over $\Hlin$ and the threshold link $\psi_{1/2}$.
\end{example}

\begin{figure}
\centering
\includegraphics[width=0.8\linewidth]{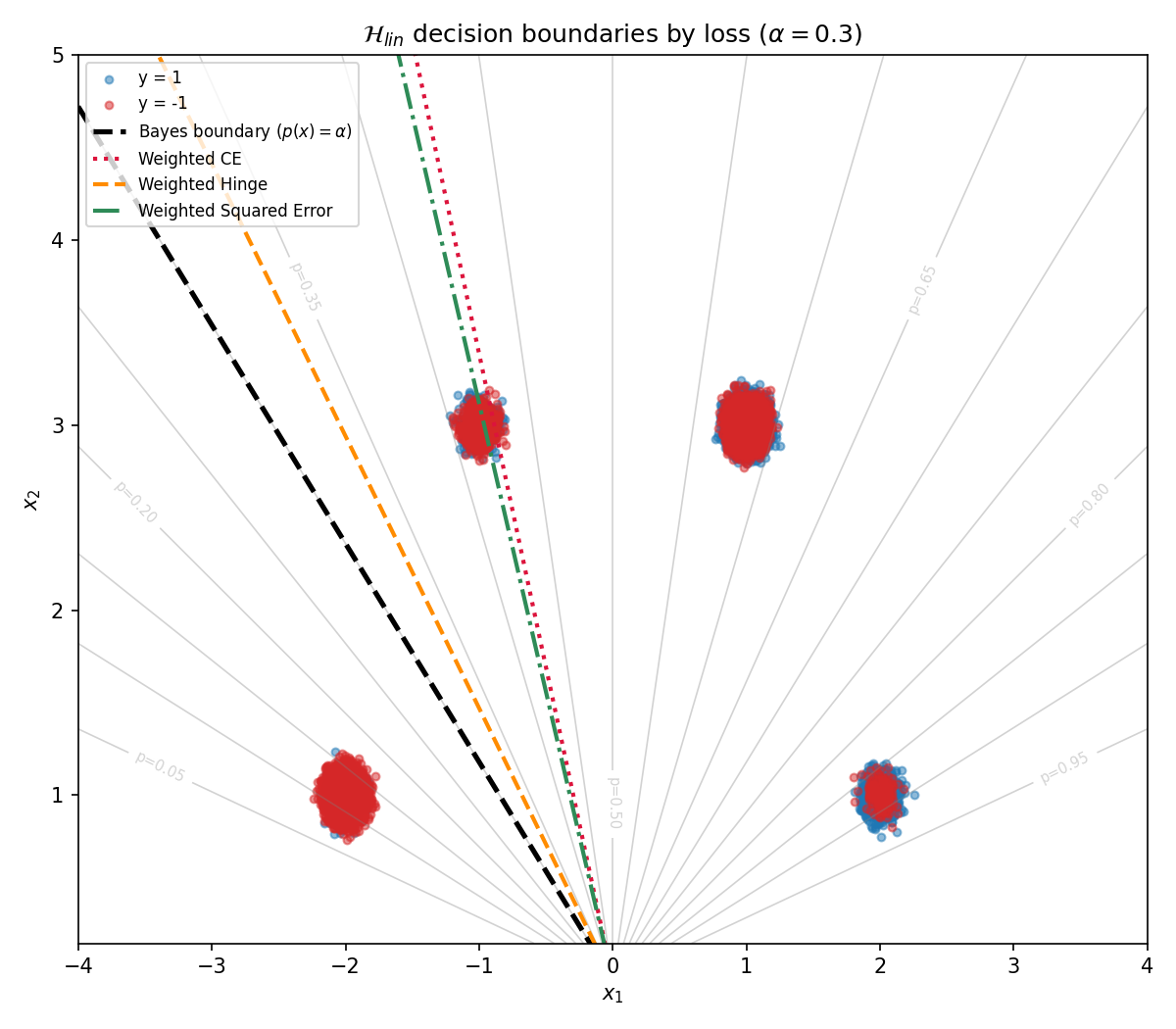}
\caption{Counterexample where weighted cross entropy and weighted squared error (over $\Hlin$) yield incorrect decisions for $\ell^\alpha$, giving a counterexample for $\Hlin$-consistency. Note that noise in the visualization of $X$ is only for visualization of the conditional label distribution.}\label{fig:antisymm-counterex}
\end{figure}

\Cref{ex:antisymm-counterex} is visualized in \Cref{fig:antisymm-counterex} with $\alpha = 0.3$, with learned boundaries for weighted hinge, cross entropy, and squared error.

Given the logit structure of $p(x)$, we can quickly verify that the boundary $\{x : p(x) = \alpha\}$ is linear for any $\alpha \in (0,1)$, though the slope of this boundary changes in $\alpha$.
As a result, we conclude that \Cref{ex:antisymm-counterex} satisfies \Cref{assum:ell-P-minimizable} with the model class $\Hlin$.

At $\alpha = 0.3$, the Bayes decisions at the four points of \Cref{ex:antisymm-counterex} are $(-1,+1,+1,+1)$, since $x_1/x_2 \in \{-2,-\tfrac 1 3, \tfrac 1 3, 2\}$ against the threshold $\log(\alpha/(1-\alpha)) \approx -0.847$, witnessed exactly by $w = (1,1) \in \Hlin$.
\Cref{assum:ell-P-minimizable} therefore holds regardless of $P_X$, since it is a condition on the support of $P$ alone.
Using a numerical solver, we find that the $L^\alpha$-population-risk minimizers under $P_X$ are $w^\ast_{XE} \approx (0.677, 0.204)$ for weighted cross entropy and $w^\ast_{sq} \approx (0.662, 0.215)$ for weighted squared error, both giving decisions $(-1,-1,+1,+1)$, disagreeing with the Bayes decisions at $x = (-1,3)$.
Weighted cross entropy is convex over $\Hlin$ unconditionally (\Cref{assum:convexity} holds), so this failure is not a non-convexity artifact.
The population risk is strictly convex here, since the four points span $\reals^2$, and its unique global minimizer $w^L \approx (0.677,0.204)$ is not $w = (1,1)$, the boundary-achieving witness, and therefore \Cref{assum:stationarity} fails.
This shows \Cref{assum:stationarity} cannot be dropped from \Cref{thm:weighted-analogs-H-consistency}, even granting \Cref{assum:convexity} and \Cref{assum:ell-P-minimizable}.

\section{Additional Empirical Results}\label{app:additional-exp}

\noindent\textbf{Loss Functions}
\begin{itemize}[itemsep=2pt,parsep=2pt,partopsep=0pt,topsep=0pt,leftmargin=1.2em]
    \item \textbf{Cross entropy (CE):} This is the most common loss function used for classification tasks and we use it as our baseline. In addition to simply training our models on cross entropy and using the $\argmax$ link to predict the class with the highest score, we also emulate ``clever'' post-processing, computing a weighted $\argmax$ where the weights are randomly sampled, and we use the weight yielding the best performance on the validation set. 
\item \textbf{Weighted Cross entropy:} This is a common cost-sensitive modification of the cross entropy loss used for classification in imbalanced data and cost-sensitive settings as studied in \S~\ref{sec:weighted-XE-binary}. The class scores are re-weighted in this loss based on the average misclassification cost from the cost matrix for that outcome. 
\item \textbf{Embeddings (\citet{finocchiaro_embedding_2024}):} This is the cost-sensitive surrogate loss function we analyze in~\cref{sec:embeddings-H-consistent}. It uses the cost matrix to create a piecewise linear and convex loss function that generalizes the hinge loss to cost-sensitive classification. In addition to the standard version of the loss, we also use a `softmax' version of the loss function in which the scores of different classes are not directly predicted. Instead, the predicted point is parameterized as a convex combination of ``embedding points,'' one for each class. In this version, we predict the coefficients of the convex combination. 
\end{itemize}

\subsection{Description of datasets}
\begin{itemize}[itemsep=2pt,topsep=0pt,partopsep=0pt,parsep=2pt,leftmargin=1.2em]
    \item \textbf{Synthetic:} This is the task from \Cref{sec:XE-not-H-consistent} with $\alpha = \frac{1}{6}$.
    \item \textbf{German Credit~\cite{statlog_144}:} The goal is to differentiate between those with ``good'' and ``bad'' credit risks based on 20 observed attributes like credit history and requested loan amount. They also specify a cost matrix of $\ell = \begin{bmatrix}0&5\\1& 0\end{bmatrix}$. Additionally, we consider a variant in which one can choose to defer by paying a constant cost of $\frac{1}{2}$ units.
    \item \textbf{Student Performance~\cite{valentim21predict}:} The goal is to predict whether a student is going to either (a) drop out, (b) continued enrollment (e.g., $>$ 4 years), or (c) graduate from their undergraduate degree at the end of the expected duration of the course. There are 36 features, such as program choice and past academic performance. We use the cost matrix $\ell = \begin{bmatrix}0&3& 5\\1& 0& 3\\ 3& 1& 0\end{bmatrix}$.
    \item \textbf{Diabetes~\cite{teboul22diabetes}:} The goal is to predict whether an individual either (a) has no diabetes, (b) is pre-diabetic, or (c) has diabetes based on 21 lifestyle features like physical activity level and history of certain medical conditions like high blood pressure. We use the same cost matrix as student performance.
\end{itemize}

\begin{table*}[h]
    \centering
    \begin{tabular}{clcc}
    \toprule
    Dataset & Loss & Cost-Sensitive Loss & 0-1 Accuracy\\
    \midrule
    \multirow{5}{*}{\makecell{Synthetic\\\textcolor{black!40}{(10000 Samples)}}} & Cross Entropy & 0.427 ± 0.004 & \textbf{0.829 ± 0.001} \\
    & Cross Entropy + Post-processing & 0.391 ± 0.004 & 0.807 ± 0.002 \\
    \arrayrulecolor{black!40} \cmidrule{2-4} \arrayrulecolor{black}
& Embeddings & \textbf{0.304 ± 0.001} & 0.767 ± 0.001 \\
    & Weighted Cross Entropy & 0.313 ± 0.002 & 0.741 ± 0.002 \\
    \midrule
    \multirow{5}{*}{\makecell{German\\Credit\\\textcolor{black!40}{(1000 Samples)}}} & Cross Entropy & 0.756 ± 0.016 & \textbf{0.720 ± 0.004} \\
    & Cross Entropy + Post-processing & 0.626 ± 0.012 & 0.619 ± 0.009 \\
    \arrayrulecolor{black!40} \cmidrule{2-4} \arrayrulecolor{black}
& Embeddings & \textbf{0.572 ± 0.007} & 0.592 ± 0.004 \\
    & Weighted Cross Entropy & 0.573 ± 0.005 & 0.515 ± 0.007 \\
    \midrule
    \multirow{4}{*}{\makecell{German\\Credit\\w/ Deferral}} & Cross Entropy & 0.682 ± 0.015 & - \\
    \arrayrulecolor{black!40} \cmidrule{2-4} \arrayrulecolor{black}
& Embeddings & \textbf{0.492 ± 0.006} & - \\
    & Weighted Cross Entropy & 0.513 ± 0.006 & - \\
    \midrule
    \multirow{4}{*}{\makecell{Student\\Performance\\\textcolor{black!40}{(4424 Samples)}}} & Cross Entropy & 0.639 ± 0.006 & \textbf{0.752 ± 0.002} \\
    & Cross Entropy + Post-processing & 0.624 ± 0.006 & 0.730 ± 0.002 \\
    \arrayrulecolor{black!40} \cmidrule{2-4} \arrayrulecolor{black}
& Embeddings & \textbf{0.584 ± 0.004} & 0.699 ± 0.001 \\
    & Weighted Cross Entropy & 0.637 ± 0.006 & 0.737 ± 0.002 \\
    \midrule
    \multirow{4}{*}{\makecell{Diabetes\\\textcolor{black!40}{(253680 Samples)}}} & Cross Entropy & 0.673 ± 0.002 & \textbf{0.844 ± 0.000} \\
    & Cross Entropy + Post-processing & \textbf{0.655 ± 0.002} & 0.839 ± 0.000 \\
    \arrayrulecolor{black!40} \cmidrule{2-4} \arrayrulecolor{black}
& Embeddings & 0.675 ± 0.002 & 0.783 ± 0.004 \\
    & Weighted Cross Entropy & \textbf{0.660 ± 0.002} & 0.817 ± 0.001 \\
    \bottomrule
    \end{tabular}
    \caption{\textbf{Ablations: More Samples}. The table above summarizes the performance of linear models trained on different loss functions, datasets and metrics. The results are presented in the form of (Expected Value $\pm$ Standard Error of the Mean).}
    \label{tab:results_more_data}
\end{table*}

\begin{table*}[h]
    \centering
    \begin{tabular}{clcc}
    \toprule
    Dataset & Loss & Cost-Sensitive Loss ($\downarrow$) & 0-1 Accuracy ($\uparrow$)\\
    \midrule
    \multirow{5}{*}{\makecell{Synthetic\\\textcolor{black!40}{(10000 Samples)}}} & Cross Entropy & 0.370 ± 0.003 & \textbf{0.828 ± 0.001} \\
    & Cross Entropy + Post-processing & 0.331 ± 0.003 & 0.811 ± 0.001 \\
    \arrayrulecolor{black!40} \cmidrule{2-4} \arrayrulecolor{black}
& Embeddings  & \textbf{0.303 ± 0.002} & 0.775 ± 0.001 \\
    & Weighted Cross Entropy & 0.317 ± 0.002 & 0.722 ± 0.002 \\
    \midrule
    \multirow{5}{*}{\makecell{German\\Credit\\\textcolor{black!40}{(1000 Samples)}}} & Cross Entropy & 0.747 ± 0.012 & \textbf{0.710 ± 0.004} \\
    & Cross Entropy + Post-processing & 0.614 ± 0.011 & 0.615 ± 0.008 \\
    \arrayrulecolor{black!40} \cmidrule{2-4} \arrayrulecolor{black}
& Embeddings & 0.608 ± 0.008 & 0.552 ± 0.014 \\
    & Weighted Cross Entropy & \textbf{0.602 ± 0.008} & 0.465 ± 0.012 \\
    \midrule
    \multirow{4}{*}{\makecell{German\\Credit\\w/ Deferral}} & Cross Entropy & 0.655 ± 0.013 & - \\
    \arrayrulecolor{black!40} \cmidrule{2-4} \arrayrulecolor{black}
& Embeddings & \textbf{0.493 ± 0.004} & - \\
    & Weighted Cross Entropy & 0.507 ± 0.008 & - \\
    \midrule
    \multirow{4}{*}{\makecell{Student\\Performance\\\textcolor{black!40}{(4424 Samples)}}} & Cross Entropy & 0.622 ± 0.004 & \textbf{0.746 ± 0.001} \\
    & Cross Entropy + Post-processing & 0.589 ± 0.004 & 0.717 ± 0.003 \\
    \arrayrulecolor{black!40} \cmidrule{2-4} \arrayrulecolor{black}
& Embeddings & \textbf{0.574 ± 0.004} & 0.718 ± 0.002 \\
    & Weighted Cross Entropy & 0.677 ± 0.004 & 0.732 ± 0.001 \\
    \midrule
    \multirow{4}{*}{\makecell{Diabetes\\\textcolor{black!40}{(253680 Samples)}}} & Cross Entropy & 0.754 ± 0.001 & \textbf{0.842 ± 0.000} \\
    & Cross Entropy + Post-processing & 0.754 ± 0.001 & \textbf{0.842 ± 0.000} \\
    \arrayrulecolor{black!40} \cmidrule{2-4} \arrayrulecolor{black}
& Embeddings & \textbf{0.652 ± 0.002} & 0.837 ± 0.000 \\
    & Weighted Cross Entropy & 0.657 ± 0.001 & 0.817 ± 0.000 \\
    \bottomrule
    \end{tabular}
    \caption{\textbf{Ablations: Larger Model}. The table above summarizes the performance of 4-layer neural network trained on different loss functions, datasets and metrics. The results are presented in the form of (Expected Value $\pm$ Standard Error of the Mean).}
    \label{tab:results_nn}
\end{table*}

\begin{table*}[h]
    \centering
    \resizebox{\textwidth}{!}{
    \begin{tabular}{clcccccc}
    \toprule
    \multirow{2}{*}{Dataset} & \multirow{2}{*}{Loss} & \multicolumn{3}{c}{Linear} & \multicolumn{3}{c}{NN} \\
    \cmidrule(lr){3-5} \cmidrule(lr){6-8}
    & & Train & Val & Test & Train & Val & Test
    \\
    \midrule
    \multirow{5}{*}{\makecell{Synthetic\\\textcolor{black!40}{(Low Overfitting)}}} & Cross Entropy & 0.430 ± 0.003 & 0.429 ± 0.004 & 0.427 ± 0.004 & 0.362 ± 0.002 & 0.366 ± 0.004 & 0.370 ± 0.003 \\
    & Cross Entropy + Post-processing & 0.396 ± 0.003 & 0.391 ± 0.004 & 0.391 ± 0.004 & 0.331 ± 0.002 & 0.327 ± 0.003 & 0.331 ± 0.003 \\
    \arrayrulecolor{black!40} \cmidrule{2-8} \arrayrulecolor{black}
& Embeddings & 0.302 ± 0.001 & 0.309 ± 0.002 & 0.304 ± 0.001 & 0.295 ± 0.001 & 0.303 ± 0.002 & 0.303 ± 0.002 \\
    & Weighted Cross Entropy & 0.314 ± 0.002 & 0.316 ± 0.002 & 0.313 ± 0.002 & 0.310 ± 0.001 & 0.318 ± 0.002 & 0.317 ± 0.002 \\
    \midrule
    \multirow{5}{*}{\makecell{German\\Credit\\\textcolor{black!40}{(High Overfitting)}}} & Cross Entropy & 0.590 ± 0.005 & 0.734 ± 0.016 & 0.756 ± 0.016 & 0.427 ± 0.011 & 0.739 ± 0.013 & 0.747 ± 0.012 \\
    & Cross Entropy + Post-processing & 0.579 ± 0.004 & 0.570 ± 0.009 & 0.626 ± 0.012 & 0.470 ± 0.008 & 0.549 ± 0.008 & 0.614 ± 0.011 \\
    \arrayrulecolor{black!40} \cmidrule{2-8} \arrayrulecolor{black}
& Embeddings & 0.381 ± 0.003 & 0.574 ± 0.007 & 0.572 ± 0.007 & 0.383 ± 0.017 & 0.573 ± 0.008 & 0.608 ± 0.008 \\
    & Weighted Cross Entropy & 0.473 ± 0.006 & 0.555 ± 0.011 & 0.573 ± 0.005 & 0.499 ± 0.014 & 0.587 ± 0.009 & 0.602 ± 0.008 \\
    \midrule
    \multirow{4}{*}{\makecell{German\\Credit\\w/ Deferral}} & Cross Entropy & 0.543 ± 0.005 & 0.676 ± 0.015 & 0.682 ± 0.015 & 0.382 ± 0.008 & 0.642 ± 0.012 & 0.655 ± 0.013 \\
    \arrayrulecolor{black!40} \cmidrule{2-8} \arrayrulecolor{black}
& Embeddings & 0.333 ± 0.003 & 0.488 ± 0.005 & 0.492 ± 0.006 & 0.427 ± 0.009 & 0.469 ± 0.004 & 0.493 ± 0.004 \\
    & Weighted Cross Entropy & 0.411 ± 0.002 & 0.509 ± 0.007 & 0.513 ± 0.006 & 0.372 ± 0.010 & 0.501 ± 0.007 & 0.507 ± 0.008 \\
    \midrule
    \multirow{4}{*}{\makecell{Student\\Performance\\\textcolor{black!40}{(High Overfitting)}}} & Cross Entropy & 0.546 ± 0.002 & 0.614 ± 0.006 & 0.639 ± 0.006 & 0.436 ± 0.004 & 0.613 ± 0.006 & 0.622 ± 0.004 \\
    & Cross Entropy + Post-processing & 0.565 ± 0.003 & 0.575 ± 0.006 & 0.624 ± 0.006 & 0.458 ± 0.003 & 0.566 ± 0.005 & 0.589 ± 0.004 \\
    \arrayrulecolor{black!40} \cmidrule{2-8} \arrayrulecolor{black}
& Embeddings & 0.461 ± 0.001 & 0.569 ± 0.004 & 0.584 ± 0.004 & 0.341 ± 0.002 & 0.566 ± 0.005 & 0.574 ± 0.004 \\
    & Weighted Cross Entropy & 0.542 ± 0.002 & 0.615 ± 0.006 & 0.637 ± 0.006 & 0.522 ± 0.005 & 0.661 ± 0.006 & 0.677 ± 0.004 \\
    \midrule
    \multirow{4}{*}{\makecell{Diabetes\\\textcolor{black!40}{(Low Overfitting)}}} & Cross Entropy & 0.668 ± 0.001 & 0.672 ± 0.002 & 0.673 ± 0.002 & 0.752 ± 0.000 & 0.751 ± 0.001 & 0.754 ± 0.001 \\
    & Cross Entropy + Post-processing & 0.652 ± 0.001 & 0.652 ± 0.001 & 0.655 ± 0.002 & 0.752 ± 0.000 & 0.751 ± 0.001 & 0.754 ± 0.001 \\
    \arrayrulecolor{black!40} \cmidrule{2-8} \arrayrulecolor{black}
& Embeddings & 0.664 ± 0.002 & 0.672 ± 0.003 & 0.675 ± 0.002 & 0.614 ± 0.002 & 0.649 ± 0.002 & 0.652 ± 0.002 \\
    & Weighted Cross Entropy & 0.656 ± 0.001 & 0.658 ± 0.001 & 0.660 ± 0.002 & 0.639 ± 0.000 & 0.654 ± 0.001 & 0.657 ± 0.001 \\
    \bottomrule
    \end{tabular}}
    \caption{Overfitting in Linear Models vs. 4-Layer Neural Networks.}
    \label{tab:results_overfitting}
\end{table*}

All experiments were conducted on a personal laptop using macOS on CPUs and results were returned on the order of hours.
In implementing the embeddings surrogates, we pre-trained a separate (linear) model which approximates the surrogate loss with high precision, and use this model to compute loss values; this pre-training process took on the order of minutes.

\end{document}